\documentclass{article}

\usepackage{PRIMEarxiv}
\usepackage[numbers]{natbib}
\usepackage{amsmath}
\usepackage{doi}

\usepackage[utf8]{inputenc} 
\usepackage[T1]{fontenc}    
\usepackage{newunicodechar}
\newunicodechar{̈}{"{}}
\usepackage{hyperref}       
\usepackage{url}            
\usepackage{booktabs}       
\usepackage{amsfonts}       
\usepackage{nicefrac}       
\usepackage{microtype}      
\usepackage{lipsum}
\usepackage{fancyhdr}       
\usepackage{graphicx}       

\usepackage{array}
\usepackage[table,xcdraw]{xcolor}
\usepackage{adjustbox}
\graphicspath{{media/}}     

\title{Balancing Accuracy, Calibration, and Efficiency in Active Learning with Vision Transformers Under Label Noise
}

\author{%
  Moseli Mots’oehli$^{1}$\quad Hope Mogale$^{2}$\quad Kyungim Baek$^{1}$\\[1ex]
  $^{1}$Department of Information and Computer Sciences, University of Hawai‘i at Manoa, Honolulu, HI, USA\\
  $^{2}$Department of Computer Science, University of Pretoria, Pretoria, South Africa\\[0.5ex]
  \texttt{\{moselim, kyungim\}@hawaii.edu}, \ \texttt{hope.mogale@up.ac.za}
}

\begin{document}
\maketitle

\begin{abstract}
Fine-tuning pre-trained convolutional neural networks on ImageNet for downstream tasks is well-established. Still, the impact of model size on the performance of vision transformers in similar scenarios, particularly under label noise, remains largely unexplored. Given the utility and versatility of transformer architectures, this study investigates their practicality under low-budget constraints and noisy labels. We explore how classification accuracy and calibration are affected by symmetric label noise in active learning settings, evaluating four vision transformer configurations (Base and Large with 16x16 and 32x32 patch sizes) and three Swin Transformer configurations (Tiny, Small, and Base) on CIFAR10 and CIFAR100 datasets, under varying label noise rates. Our findings show that larger ViT models (ViTl32 in particular) consistently outperform their smaller counterparts in both accuracy and calibration, even under moderate to high label noise, while Swin Transformers exhibit weaker robustness across all noise levels. We find that smaller patch sizes do not always lead to better performance, as ViTl16 performs consistently worse than ViTl32 while incurring a higher computational cost. We also find that information-based Active Learning strategies only provide meaningful accuracy improvements at moderate label noise rates, but they result in poorer calibration compared to models trained on randomly acquired labels, especially at high label noise rates. We hope these insights provide actionable guidance for practitioners looking to deploy vision transformers in resource-constrained environments, where balancing model complexity, label noise, and compute efficiency is critical in model fine-tuning or distillation.
\end{abstract}


\section{Introduction}\label{sec:Introduction}
Transformer-based models have achieved great success in multiple vision \cite{wang:onepeace23,Srivastava:OmniVecLR23,Zong:DETRs23,kirillov:SAM23} and language tasks \cite{Borgeaud:ImprovingLM21}, and power the most commonly used generative image, text, and video products. Despite this success, the Vision Transformer (ViT)'s adoption and the exploration of their properties in application domains such as agriculture, remote sensing, Disaster management, and more specialized areas of Deep Learning (DL), such as Deep Active Learning (DAL), and learning with label noise, have been relatively slower. Given a large collection of unlabeled images, a limited labeling budget, and a DL model, DAL seeks to strategically sample fewer images for labeling in such a way they lead to the optimal DL model generalization performance within the labeling budget. However, the provided labels by the labeling oracle may not always be correct, leading to complexities in learning stable decision boundaries for the DL model. 

The majority of the work done in DAL, and DAL under label noise uses Convolutional Neural Networks (CNN)s \cite{LeCun:CNN98}, and focuses on the design of noise-robust DAL query strategies that outperform baseline random, and entropy-based methods on pre-defined image classification task \cite{Ren:DALSurvey20, Cordeiro:DeepNoisyAnnotations20,Mots'oehli:DeepActiveLabelNoise23}. For this reason, the influence of the underlying DL model is often neglected, thus leading to multiple studies comparing and reporting results on DAL strategies under varying DL model architectures, number of parameters, pre-training datasets, and training configurations on a downstream DAL task using the same dataset. This oversight, recently also investigated in \cite{bajracharya:interdependence24} using only CNNs, can result in performance and robustness gains due to pre-training, model architecture type, or size being mistakenly attributed to a new proposed DAL strategy or robust loss function for label noise. Moreover, studies under standardized conditions have shown that without extensive hyper-parameter tuning, most DAL strategies perform no better than random query or entropy-based selection on some datasets \cite{li:empiricalEfficacy22,Mots'oehli:DeepActiveLabelNoise23,gashi:DALRealityCheck24}. Building on these findings and results from \cite{Mots'oehli:GCIViTAL24}, demonstrating that all else being equal, ViTs outperform CNNs considerably in DAL under label noise on CIFAR10, CIFAR100, the Food101 dataset, and the chest x-ray images (pneumonia) dataset, this paper addresses the question: "What is the impact of label noise on different ViT model sizes and capacity for image classification in the active learning setting"?

\begin{figure}[htbp]
    \centering
    \includegraphics[width=\textwidth]{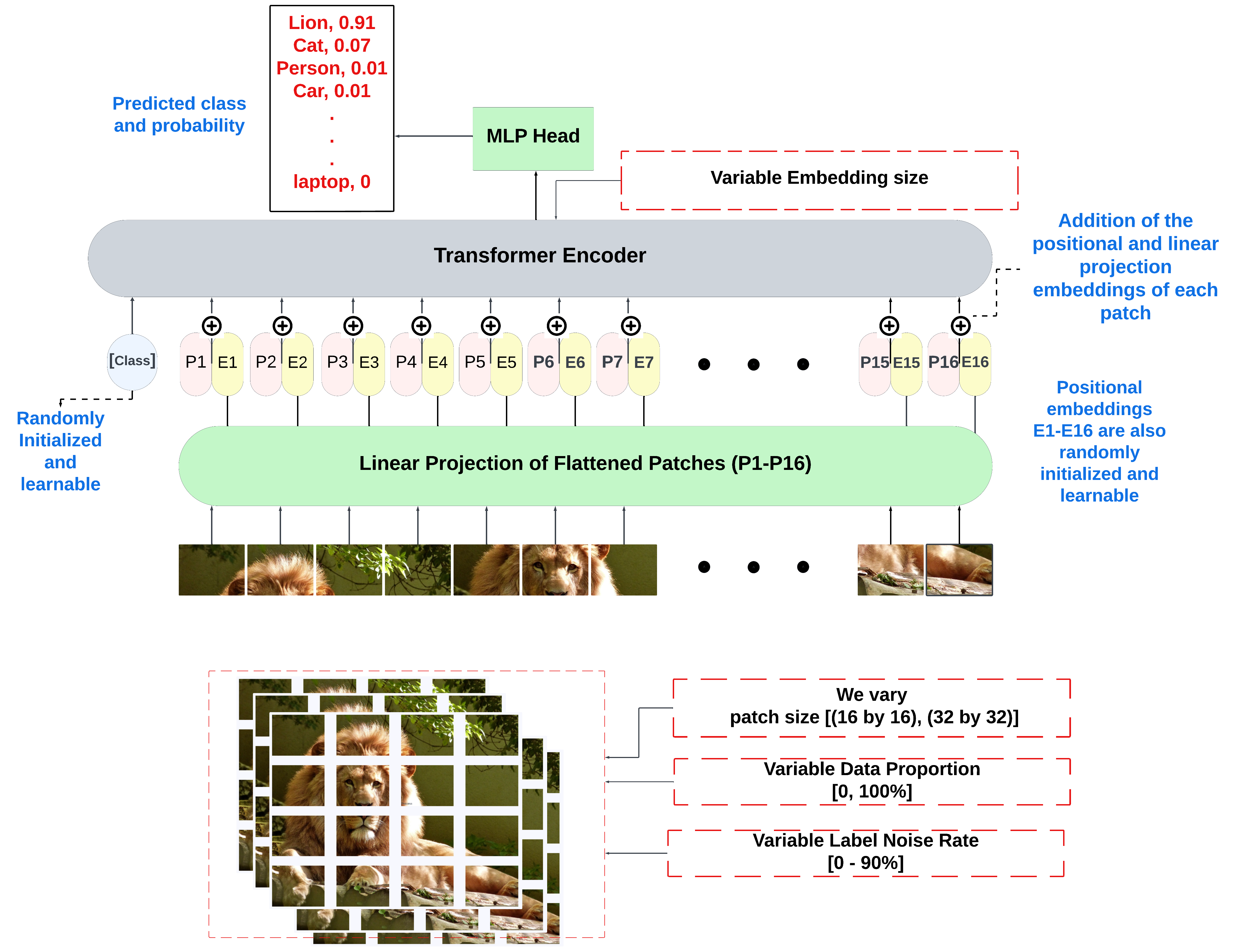} 
    \caption{The key components involved in fine-tuning a transformer under label noise. The aspects we vary in our experiments are indicated in red. The Active Learning variation spans the entire diagram. [Adapted from \cite{Mots'oehli:GCIViTAL24}]}
    \label{fig:all}
\end{figure}

This study investigates the relationship between ViT model size, capacity, and label noise for image classification to address this question. Figure \ref{fig:all} illustrates a vision transformer as well as the components we vary in our experiments. We explore four ViTs \cite{Kolesnikov:ViT21}, three Swin transformers \cite{Liu:swinTransformer21} that vary in the number of parameters (size), and the dimensions of the patch embeddings used for feature extraction (capacity). We hypothesize that larger ViTs in terms of embedding size and number of tokens (smaller patch size means more tokens), with their enhanced ability to learn intricate representations, while more computationally expensive, will show greater robustness to label noise during the DAL cycle. We also hypothesize that neither size nor capacity will make a significant difference at very high label noise rates. We evaluate the ViTs under known and controlled levels of label noise, monitoring model test accuracy and calibration using the Brier score \cite{Brier:VERIFICATIONOF50}, throughout the experiment to assess how the accuracy and calibration of models of different sizes and capacities are affected by increasing label noise. To ensure the effect of ViT input patch size and capacity on label noise is clear, we use only a small selection of DAL query strategies:(random sampling, entropy sampling, and GCI\_ViTAL \cite{Mots'oehli:GCIViTAL24}) to minimize any unexplainable influences of individual DAL strategies. We train and evaluate the models on the commonly used CIFAR10 and CIFAR100 classification datasets.

\textbf{Contribution:} The main contributions of this study can be summarized as follows:

\begin{itemize}
    \item By investigating the impact of varying symmetric label noise rates on the generalization performance of ViT model architectures of varying input patch sizes and capacity in the DAL setting, we fill the gap in the existing literature that often focuses on the design of DAL strategies irrespective of advances in DL model architectures and their properties.
    \item We experimentally show that selecting the largest, highest-capacity model is not always the best fine-tuning strategy when labeling budgets are low, label noise is present, and computational resources are constrained. By challenging the common practice of defaulting to base or larger models, our findings highlight the need for data- and situation-specific solutions. This offers practitioners a more realistic framework for deciding which transformer model and size to choose under real-world conditions.
\end{itemize}

\section{Related Work}\label{sec:RelatedWork}
In this section, we review the existing literature on image classification in noisy settings, focusing on the intersection between patch size, embedding dimensions, and DAL strategies with ViT-based models.

\subsection{Image Classification}\label{textbfsubsec:Label_noise_classification}
Image classification has been a key task in computer vision for many years. Significant advancements have been made through CNN-based deep learning models like AlexNet \cite{Krizhevsky:ImageNetCW2012}, InceptionNet \cite{Szegedy:inceptionNet15}, and ResNet \cite{He:ResNet16}, as well as large labeled datasets such as CIFAR10, CIFAR100 \cite{krizhevsky:Cifar09}, and ImageNet \cite{Deng:ImageNet09}. However, since not everyone has the necessary computational resources for training all layers of Deep Neural networks (DNNs), or the budget to label large amounts of image data, the field has shifted toward transfer learning. This works by downloading models trained on large, freely accessible datasets and then fine-tuning them on downstream tasks using smaller labeled datasets and computing resources \cite{yosinski:transferable14}. The current state of the art in image classification relies on self-supervised learning \cite{chen:SimCLR20, he:momentumContrast20} from large labeled or unlabeled datasets using transformer-based models such as DeiT \cite{touvron:DieT21}, which treat images as sequences of patches and utilize self-attention mechanisms to learn robust image representations. Once these models are trained, a simple multilayer perceptron classification head is added to map the final layer representations to image classes, requiring significantly less labeled domain-specific data. 

\subsection{Vision Transformers for Active Image Classification}\label{sec:ViTClassifcation}
Previous works that adopt the ViTs for image classification in the DAL domain include \cite{caramalau:visualTaskAware21} and \cite{HE:ViTMedical21}. In \cite{caramalau:visualTaskAware21}, the authors introduce a novel DAL query strategy that combines CNN layers for local dependencies and ViTs to capture non-local dependencies while jointly minimizing a task-aware objective. They achieve state-of-the-art performance on most AL-based benchmarks. However, their method has scalability limitations due to ViT's large parameter space and potential batch size restrictions in training. \cite{HE:ViTMedical21} reaches a similar conclusion. Their work demonstrates that, while ViTs produce informative and task-aware DAL queries on CIFAR10 and CIFAR100, they are considerably larger than CNNs in terms of model parameters for them to be a viable replacement in DAL with the existing hardware and ease to parallelize DAL training. The work of Rotman and Reichart \cite{Rotman:MultiTaskALTransformerBased22} compares different DAL methods on different text classification datasets using transformer-based models. While their work is not focused on image classification or the vision transformer, they demonstrate that transformer-based models tend to lead to inconsistent and poor results in the DAL setting when using basic DAL strategies. They show that query selection based on a transformer learner sometimes leads to the selection of clusters of neighboring outliers that destabilize training.

\subsection{Patch Size, Embedding Size and Performance}\label{textbfsubsec:patch_and_embedding_size}
The choice of patch and embedding size in ViTs plays an important role in model performance, especially in real-world datasets that are not always as big as we would like them to be, or we can not always guarantee the correctness of the labels. ViT patch size affects image resolution processing: smaller patches allow for finer feature extraction. In comparison, larger patches reduce computational complexity at the expense of detail required for optimal performance \cite{Zhai:ScalingVT21,caron:emergingProps21}. On the other hand, the ViT embedding size affects the model's capacity to learn complex functions, and thus its ability to generalize from the available training data \cite{Chen:rossViTCM21,Liu:swinTransformer21}. However, this is not always the case in less ideal situations such as training on small datasets \cite{Shao:TransMeetSmall22,Zhu:UnderstandingWV23}, and in particular, \cite{Zhai:ScalingVT21} find that a small dataset size bottlenecks the benefits of scaling compute and model size since there is only so much you can learn from low entropy.  

 Despite the insights and progress, a notable gap exists in exploring how patch and embedding size interact with label noise for image classification using ViTs. This is particularly important in DAL settings, where computational efficiency, optimizing performance, and calibration on a small labeling budget are more important and realistic than unconstrained absolute model performance. The next section describes the models used, datasets, active learning algorithms, and label noise.

\section{Methodology}\label{sec:Methodology}
In this section, we describe the two transformer architecture variants used in the experiments and the different model sizes per architecture. We briefly discuss the label noise injection process, active learning query strategies, and datasets used.

\subsection{ViT Model Size and Capacity}\label{subsec:ViT_Size_Capacity_Meth}

\textbf{ViT: } We use the original transformer implementation \cite{Kolesnikov:ViT21} in our classification experiments, with 4 different configurations varying in patch size and embedding size.  The variants \texttt{ViTb16}, \texttt{ViTb32}, \texttt{ViTl16}, and \texttt{ViTl32} correspond to base (768-dimensional) and large (1024-dimensional) embedding sizes, combined with patch sizes of 16×16 and 32×32 pixels respectively. The base models consist of 12 transformer layers and 12 attention heads, whereas the large models scale up to 24 layers and 16 attention heads. Table \ref{table:models} summarizes the details above for clarity.                                                                                                    
\textbf{SwinV2: }The SwinV2 transformer models \cite{Liu:swinTransformer21} utilize a hierarchical architecture and overlapping shifted windows for patch processing (as opposed to the non-overlapping used by ViT), enhancing their ability to model both local and global features effectively. We include three variants, all with a $4 \times 4$ pixel patch size. The \texttt{SwinV2t} and \texttt{SwinV2s} models have an embedding size of 768 and differ in the number of transformer layers, with 12 and 24 transformer layers respectively. The primary choice, why SwinV2 was highly favored, is because, in SwinV1, the embedding size \( E \) is defined as a function of the number of input channels \( C \) and the depth \( D \) of the model, typically represented as:

\begin{equation}
E_{\text{V1}} = f(C, D)
\end{equation}

In contrast, SwinV2 introduces a more flexible embedding mechanism that allows for variable embedding sizes across different layers, enhancing its ability to capture complex patterns. This can be mathematically expressed as:

\begin{equation}
E_{\text{V2}} = g(C, D, R)
\end{equation}

where \( R \) represents the resolution of the input image, which allows for an adaptive scaling based on input characteristics. 
The windowing mechanism is another key area that demonstrates one of the key differences between the two. SwinV1 utilizes a fixed window size \( W \) for self-attention calculations, typically set to \( W = 7 \times 7 \). The self-attention complexity is thus quadratic for the number of tokens \( N \):

\begin{equation}
\mathcal{O}(N^2)
\end{equation}

The self-attention mechanism operates by computing all token pairs within each context window. The attention scores are calculated as follows:

\begin{equation}
A_{ij} = \frac{\exp(Q_i K_j^T / \sqrt{d_k})}{\sum_{k=1}^{N} \exp(Q_i K_k^T / \sqrt{d_k})}
\end{equation}

where: \( A_{ij} \) is the attention score between token \( i \) and token \( j \), \( Q_i \) is the query vector for token \( i \), \( K_j \) is the key vector for token \( j \), \( d_k \) is the dimensionality of the key vectors, and \( N \) is the total number of tokens in the window. The quadratic complexity that comes with SwinV1 has drawbacks \cite{Liu:swinTransformer21} which are improved upon using the adaptive scaling in SwinV2, allowing for more efficient computations based on varying resolutions. The complexity can be expressed as:

\begin{equation}
\mathcal{O}(N d + S(R))
\end{equation}

where \( S(R) = M N d / W^2 = k H W N d / W^2 \), indicating that as resolution increases, computational efficiency improves through better utilization of local attention mechanisms. 

This results in a more efficient self-attention mechanism with linear complexity:

\begin{equation}
\mathcal{O}(N \cdot W'^2)
\end{equation}

This shift reduces computational overhead and allows for better feature extraction by enabling interactions between neighboring windows. In a nutshell, the \texttt{SwinV2b} model has an embedding size of 1024, 4 attention heads, and maintains 24 transformer layers. We use these ViT and SwinV2 model configurations in all our classification experiments for all datasets, label noise rates, and DAL strategies. The multi-head attention mechanism, which can be mathematically expressed as:

\begin{equation}
MultiHead(X_q, X_k, X_v) = W^O \cdot \bigg( \bigoplus_{i=1}^{h} Attention_i(X_q, X_k, X_v) \bigg) 
\end{equation}

where: \(X_q \in \mathbb{R}^{d_q}\) is the query matrix, \(X_k \in \mathbb{R}^{d_k}\) is the key matrix, \(X_v \in \mathbb{R}^{d_v}\) is the value matrix, \(W^O\) is the output projection matrix, \(h\) is the number of attention heads, and \(\bigoplus\) denotes the concatenation of the outputs from each head. Each attention head computes its output as:

\begin{equation}
Attention_i(X_q, X_k, X_v) = softmax\left(\frac{X_q W_i^Q (X_k W_i^K)^T}{\sqrt{d_k}}\right) (X_v W_i^V) 
\end{equation}

where: \(W_i^Q, W_i^K, W_i^V\) are the weight matrices for queries, keys, and values respectively. This mechanism allows the model to focus on different parts of the input sequence simultaneously, enhancing feature extraction capabilities across various scales.

\begin{table}[ht!]
\centering
\caption{Architectural details and average train times for selected SwinV2 and ViT models over CIFAR10 and CIFAR100. Abbreviations: PS = Patch Size, ES = Embedding Size, L = Layers, H = Heads, MLP = MLP Size, and $\bar{t}$ = average train time (s). The runtime values indicate the average training times across all experiments using 100\% of the training data.}
\label{table:models}
\begin{tabular}{l c c c c c c}
\toprule
\textbf{Model} & \textbf{PS} & \textbf{ES} & \textbf{L} & \textbf{H} & \textbf{MLP} & $\bar{t}$ (s) \\
\midrule
SwinV2t & 4x4  & 768  & 12 & 3  & 768  & 54 \\
SwinV2s & 4x4  & 768  & 24 & 3  & 768  & 72 \\
SwinV2b & 4x4  & 1024 & 24 & 4  & 1024 & 102 \\
\midrule
ViTb16  & 16x16 & 768  & 12 & 12 & 3072 & 85 \\
ViTb32  & 32x32 & 768  & 12 & 12 & 3072 & 41 \\
ViTl16  & 16x16 & 1024 & 24 & 16 & 4096 & 228 \\
ViTl32  & 32x32 & 1024 & 24 & 16 & 4096 & 90 \\
\bottomrule
\end{tabular}
\end{table}

\subsection{Datasets}\label{subsec:Datasets_Meth}
To investigate the impact of label noise on model patch size and capacity in DAL training, we use the widely used CIFAR10 and CIFAR100 datasets for image classification. Both datasets consist of 60,000 small, color images with a resolution of $32 \times 32$ pixels and are balanced in the number of images per class. CIFAR10 contains images from 10 common object classes, providing a straightforward classification task. CIFAR100 extends CIFAR10 with 100 classes grouped into 20 super-classes like vehicles, animals, and flowers, with finer sub-classes, offering a more fine-grained and relatively harder classification problem.

\subsection{Label Noise}\label{subsec:Label_Noise}
Since our investigation is more on how ViTs of different patch sizes, embeddings, and architecture perform in the DAL fine-tuning under label noise, we limit classification label noise to symmetric label noise and cover noise rates in the range $C_{NR} \in [0,0.9]$ with step sizes $\delta = 0.1$. Label noise is only injected during training, and the test set remains clean to measure each model's generalization performance. It is important to note that label noise is introduced only after the samples have been selected for labeling in each DAL cycle. As a result, the DAL method does not influence the symmetric label noise, and the label noise rate does not affect the DAL strategy definitively as it is unknown which samples are incorrectly labeled by the oracle.

\subsection{Deep Active Learning}\label{subsec:Active_Learning_Meth}
Below, we describe the three active learning query strategies used in our experiments. Since this work focuses on the interactions between ViT input patch size, embedding size, performance, and label noise within the active learning framework, rather than developing robust DAL strategies, we limit our experiments to a plan that is independent of the input data (random query), an Entropy-based query method, and a ViT-specific acquisition strategy -- GCI\_ViTAL. 

\paragraph{Random Query: }\label{subsubsec:randomQuery}
The random query strategy selects $k$ random samples for labeling from the unlabeled dataset. No additional information about the data, model, or task at hand is considered, and thus random query is straightforward to implement and has constant computational complexity. Despite its simplicity, random query has been shown to perform as well as most complex query strategies all else being equal \cite{Ren:DALSurvey20,li:empiricalEfficacy22,Mots'oehli:DeepActiveLabelNoise23}.

Formally, let $\mathcal{U}$ be the current unlabeled dataset, and let $D \subset \mathcal{U}$ denote the subset selected for labeling. Then:

\begin{equation}
    D = \{\, x \mid x \in \mathcal{U}, \text{ chosen uniformly at random, } |D|=K \}.
\end{equation}

\paragraph{Entropy-based Selection: }\label{subsubsec:entropy}
In image classification, this query strategy selects samples with the highest information entropy. Samples with high information entropy, where the model is less confident about the predicted class, are more informative for learning class boundaries. In selecting the samples for labeling, we first run all the unlabeled images through the model to get the predicted class probabilities for image classification, then calculate and rank the images based on entropy. We select the top $K$ unlabeled images for labeling.

Let $s_{\text{Ent}}(x)$ be the entropy score for an unlabeled sample $x \in \mathcal{U}$:

\begin{equation}
    s_{\text{Ent}}(x) 
    = 
    H\bigl(\hat{p}_\theta(y \mid x)\bigr)
    = 
    - \sum_{c=1}^{C} \hat{p}_\theta(c \mid x)\,\log \bigl[\hat{p}_\theta(c \mid x)\bigr].
\end{equation}

We then rank the unlabeled samples in descending order of $s_{\text{Ent}}(x)$ and pick the top $K$:

\begin{equation}
    D 
    = 
    \underset{x \in \mathcal{U}}{\mathrm{top}\,K}\; s_{\text{Ent}}(x).
\end{equation}

Since uncertainty-based selection depends on the model's predictions over the entire unlabeled dataset, the dataset size computationally influences this query strategy as we need to rank the samples based on uncertainty.

\begin{figure}[!ht]
	\begin{center}
		\includegraphics[width=1.0\columnwidth]{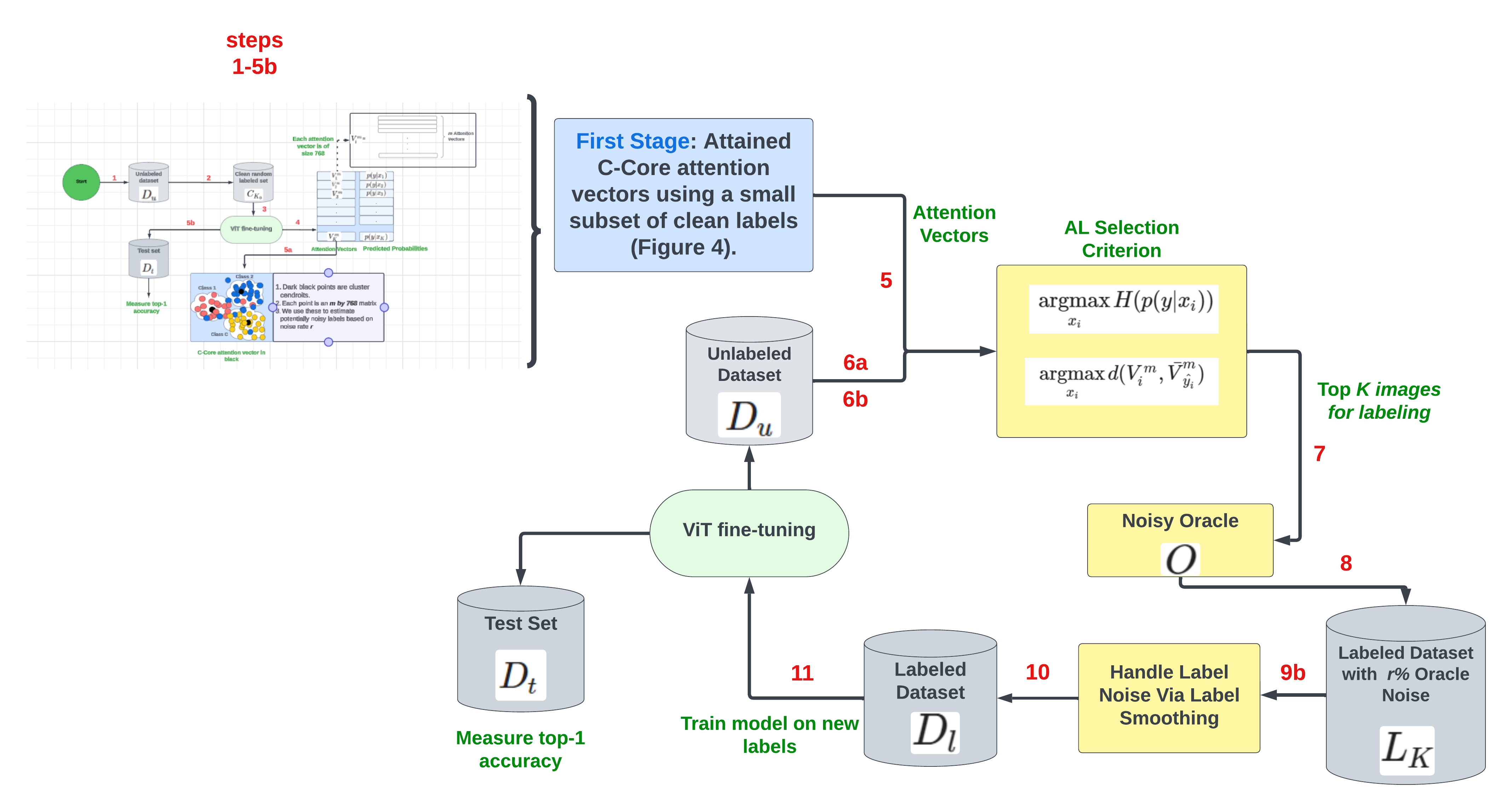}
	\end{center}
\caption{This figure shows the second stage of the GCI\_ViTAL query strategy, with C-Core attention vectors from the ViT model guiding the selection of semantically challenging samples based on their distance from class centroids. Label smoothing mitigates noise, enhancing model noise robustness \cite{Mots'oehli:GCIViTAL24}. Steps 1-5 of the strategy are implemented as described in the original paper}
\label{fig:ALFrameworkNew}
\end{figure}

\paragraph{Gradual Confidence Improvement Active Learning with Vision Transformers (GCI\_ViTAL): }\label{subsubsec:gci_vital} Designed specifically for active learning-based image classification under label noise using a ViT, this acquisition strategy combines prediction entropy and the Frobenius norm of last-layer attention vectors, comparing these vectors to a class-centric clean set used to initialize the DAL cycle. Figure \ref{fig:ALFrameworkNew}
illustrates this DAL strategy, with the combined Entropy-Frobenious norm acquisition function described in \cite{Mots'oehli:GCIViTAL24}.

\section{Experimental Setup}\label{sec:ExperimentalSetup}
This section details the dataset image transformations, the model training procedures, hardware and software configurations, and the evaluation methods for the study.

\subsection{Preprocessing}\label{subsec:preprocessing}

The datasets used in this study are CIFAR10 and CIFAR100, each adapted for Vision Transformer (ViT) and Swin Transformer architectures. Since transformers typically require larger input dimensions than the original dataset resolution, all images are resized to $224 \times 224$ pixels.

For data augmentation and normalization, the following transformations were applied:

\begin{itemize}
    \item \textit{Training Transformations:} 
    \begin{itemize}
        \item RandomResizedCrop($224$) – randomly crops and resizes the image.
        \item RandomHorizontalFlip – applies horizontal flipping with a probability of $0.5$.
        \item Convert to Tensor.
        \item Normalize with mean $[0.4914, 0.4822, 0.4465]$ and standard deviation $[0.2023, 0.1994, 0.2010]$.
        \item Resize to ($224 \times 224$).
    \end{itemize}
    \item \textit{Testing Transformations:} 
    \begin{itemize}
        \item Convert to Tensor.
        \item Resize to ($224 \times 224$).
        \item Normalize with mean $[0.4914, 0.4822, 0.4465]$ and standard deviation $[0.2023, 0.1994, 0.2010]$.
    \end{itemize}
\end{itemize}

All datasets were loaded using the PyTorch \texttt{torchvision.datasets} module. The training set was loaded with the defined augmentation pipeline, while the test set was processed with only normalization and resizing. All images were loaded in mini-batches using \texttt{torch.utils.data.DataLoader} with shuffling enabled for training data and disabled for testing. The \texttt{pin\_memory} flag was used to optimize data transfer to the GPU, and multiple workers were used to speed up data loading.

\subsection{Training, Hardware and Software}\label{subsec:training_and_hardware}
Starting with transformer models pre-trained on ImageNet-1k, we fine-tune all classifiers for 20 epochs with early stopping per DAL acquisition round, a 10-epoch tolerance, and learning rate scheduling. For each selected DAL strategy, model, and dataset, we run experiments on all noise rates in parallel on two Nvidia V100 GPUs with 24GB of RAM per GPU. The training runtimes, test accuracy, and test brier scores are recorded after each DAL selection round per DAL strategy, label noise rate, model size, and dataset. We use an initial random clean labeled training set of 1024 images to initiate the models for AL-based sample selection since DAL has a cold start problem. In each DAL round, we select the top 2048 most informative samples for annotation, selected based on the underlying DAL strategy. We use a batch size of 256 and record test performance throughout the entire labeling budget.

\subsection{Evaluation Metrics}\label{subsec:Evaluation}
We use Top-1 Accuracy, Brier Score, and Training Time in comparing the different ViT patch sizes and embedding sizes as follows:

\paragraph{Top-1 Classification Accuracy:} The standard Top-1 Classification Accuracy measures the proportion of correctly predicted samples among all test samples. Formally, for \(N\) test images, let \(\hat{y}_i\) be the predicted class for the \(i\)-th image, and \(y_i\) the ground-truth class label. Then:

\begin{equation}\label{eq:top1_accuracy}
    \text{Acc}_{\text{top-1}} = \frac{1}{N} \sum_{i=1}^{N} \mathbf{1}\{\hat{y}_i = y_i\},
\end{equation}

where \(\mathbf{1}\{\cdot\}\) is the indicator function that equals 1 if its argument is true, and 0 otherwise. A higher Top-1 Accuracy means the model more frequently places the correct class in its top prediction.

\paragraph{Brier Score: }Another proper scoring metric, it quantifies the quality of probabilistic prediction, thus measuring how well-calibrated a model is in its predictions beyond the absolute top 1-class assignment. Mathematically in the multi-classification setting, the Brier score is calculated as the weighted sum of the square differences between the predicted class probabilities and the one-hot encoded class labels over the entire test set. For a multi-classification task with $K$ classes and $N$ test samples, the Brier Score is given by:

\begin{equation}\label{eq:brier_score}
    Br = \frac{1}{N}{\sum_{i=1}^{N} \sum_{j=1}^{K} (\hat{y}_{ij} - y_{ij})^2}
\end{equation}

where $\hat{y}_{ij}$ is the predicted probability that the $i_{th}$ image belongs to class $j$, and $y_{ij}$ corresponding one-hot indicator (1 if sample $i$ is in class $j$, and 0 otherwise). A low Brier Score indicates the model is confident when it is correct and reflects uncertainty accurately for samples it is unsure about. A high Brier Score reflects a badly calibrated model.

\paragraph{Training Times:} To compare the different model configurations under label noise in different DAL acquisitions, we measure the training time per model under different datasets, label noise rates, and DAL strategies. We only measure training times per DAL cycle since all else being equal, it tells us exactly what the computational cost of using one model over the other is.

\section{Results}\label{sec:Results}
In this section, we present our findings on comparing ViT models that vary in patch and embedding sizes on CIFAR10 and CIFAR100, focusing on (i) classification accuracy under varying label noise rates, (ii) model calibration under label noise, (iii) training efficiency of the various models under different labeled data proportions, and the interplay between classification accuracy and model calibration. We look at all the results in these dimensions under the three DAL acquisition strategies (random, entropy, and GCI\_ViTAL). We include additional supporting results in Appendix \ref{sec:appendix_a_results}

\subsection{Accuracy and Label Noise} 
\begin{table}[ht]
\centering
\footnotesize
\caption{Top-1 Accuracy (\%) on CIFAR10 for different models under varying label noise rates. The color gradient highlights performance differences (green for higher accuracy, red for lower), showing how increasing label noise reduces model accuracy. The table shows that Larger ViTs (vitl16, vitl32) and SwinV2 (swinV2b, swinV2s) models generally maintain higher accuracy than smaller models under higher noise levels.}
\begin{tabular}{|l|*{10}{c|}}
\hline
\textbf{Model} & \multicolumn{10}{c|}{\textbf{Label Noise Rate}} \\
\cline{2-11}
               & \textbf{0} & \textbf{0.1} & \textbf{0.2} & \textbf{0.3} & \textbf{0.4} & \textbf{0.5} & \textbf{0.6} & \textbf{0.7} & \textbf{0.8} & \textbf{0.9} \\
\hline
\textbf{vitl16}  
& \cellcolor[HTML]{6CC17C}94.21\% 
& \cellcolor[HTML]{76C47D}93.70\% 
& \cellcolor[HTML]{83C87D}93.08\% 
& \cellcolor[HTML]{8ECB7E}92.52\% 
& \cellcolor[HTML]{A2D07F}91.55\% 
& \cellcolor[HTML]{B4D680}90.65\% 
& \cellcolor[HTML]{D5DF82}89.03\% 
& \cellcolor[HTML]{FEE683}86.18\% 
& \cellcolor[HTML]{FDC77D}81.25\% 
& \cellcolor[HTML]{F86F6C}66.96\% \\
\hline
\textbf{vitl32}  
& \cellcolor[HTML]{63BE7B}94.63\% 
& \cellcolor[HTML]{70C27C}94.02\% 
& \cellcolor[HTML]{76C47D}93.70\% 
& \cellcolor[HTML]{82C77D}93.13\% 
& \cellcolor[HTML]{8BCA7E}92.67\% 
& \cellcolor[HTML]{9DCF7F}91.79\% 
& \cellcolor[HTML]{B7D780}90.51\% 
& \cellcolor[HTML]{E4E483}88.28\% 
& \cellcolor[HTML]{FED980}84.09\% 
& \cellcolor[HTML]{FA9373}72.78\% \\
\hline
\textbf{vitb16}  
& \cellcolor[HTML]{8DCA7E}92.60\% 
& \cellcolor[HTML]{9FD07F}91.70\% 
& \cellcolor[HTML]{B2D580}90.75\% 
& \cellcolor[HTML]{B8D780}90.44\% 
& \cellcolor[HTML]{D0DE82}89.24\% 
& \cellcolor[HTML]{E6E483}88.19\% 
& \cellcolor[HTML]{FFEB84}86.94\% 
& \cellcolor[HTML]{FEDB81}84.40\% 
& \cellcolor[HTML]{FCB77A}78.54\% 
& \cellcolor[HTML]{F87B6E}68.92\% \\
\hline
\textbf{vitb32}  
& \cellcolor[HTML]{E4E383}88.29\% 
& \cellcolor[HTML]{EEE784}87.76\% 
& \cellcolor[HTML]{FEE983}86.67\% 
& \cellcolor[HTML]{FEE783}86.30\% 
& \cellcolor[HTML]{FEE182}85.40\% 
& \cellcolor[HTML]{FEDA80}84.18\% 
& \cellcolor[HTML]{FDCE7E}82.25\% 
& \cellcolor[HTML]{FDC77D}81.11\% 
& \cellcolor[HTML]{FBAC78}76.90\% 
& \cellcolor[HTML]{F86D6B}66.71\% \\
\hline
\hline
\textbf{swinV2b} 
& \cellcolor[HTML]{BFD981}90.08\% 
& \cellcolor[HTML]{C4DA81}89.87\% 
& \cellcolor[HTML]{C8DC81}89.64\% 
& \cellcolor[HTML]{D7E082}88.93\% 
& \cellcolor[HTML]{E1E383}88.40\% 
& \cellcolor[HTML]{ECE683}87.88\% 
& \cellcolor[HTML]{FEEA83}86.87\% 
& \cellcolor[HTML]{FEE282}85.60\% 
& \cellcolor[HTML]{FDCE7E}82.28\% 
& \cellcolor[HTML]{FA9773}73.37\% \\
\hline
\textbf{swinV2s} 
& \cellcolor[HTML]{CDDD82}89.38\% 
& \cellcolor[HTML]{D5DF82}89.02\% 
& \cellcolor[HTML]{DBE182}88.70\% 
& \cellcolor[HTML]{E5E483}88.22\% 
& \cellcolor[HTML]{EEE683}87.78\% 
& \cellcolor[HTML]{FEEA83}86.87\% 
& \cellcolor[HTML]{FEE683}86.11\% 
& \cellcolor[HTML]{FEDA80}84.24\% 
& \cellcolor[HTML]{FCC07B}80.04\% 
& \cellcolor[HTML]{FA9573}73.09\% \\
\hline
\textbf{swinV2t} 
& \cellcolor[HTML]{FEDE81}84.86\% 
& \cellcolor[HTML]{FEDC81}84.56\% 
& \cellcolor[HTML]{FEDB81}84.42\% 
& \cellcolor[HTML]{FDD780}83.81\% 
& \cellcolor[HTML]{FDD27F}82.96\% 
& \cellcolor[HTML]{FDCC7E}82.04\% 
& \cellcolor[HTML]{FDC67C}80.96\% 
& \cellcolor[HTML]{FCBB7A}79.25\% 
& \cellcolor[HTML]{FBA276}75.22\% 
& \cellcolor[HTML]{F8696B}65.94\% \\
\hline
\end{tabular}
\label{tab:top1_accuracy_cifar10}
\end{table}
 
\begin{table}[ht]
\centering
\footnotesize
\caption{Top-1 Accuracy (\%) by Model and Label Noise Rate across all DAL strategies and labeled data proportions of CIFAR100. We see that larger ViT and SwinV2 models generally maintain higher accuracy than smaller models under higher noise levels.}
\begin{tabular}{|l|*{10}{c|}}
\hline
\textbf{Model} & \multicolumn{10}{c|}{\textbf{Label Noise Rate}} \\
\cline{2-11}
               & \textbf{0} & \textbf{0.1} & \textbf{0.2} & \textbf{0.3} & \textbf{0.4} & \textbf{0.5} & \textbf{0.6} & \textbf{0.7} & \textbf{0.8} & \textbf{0.9} \\
\hline
\textbf{vitl16}  & \cellcolor[HTML]{69C07C}75.86\% & \cellcolor[HTML]{7AC57D}74.18\% & \cellcolor[HTML]{8ACA7E}72.64\% & \cellcolor[HTML]{99CE7F}71.10\% & \cellcolor[HTML]{ABD380}69.33\% & \cellcolor[HTML]{BED981}67.43\% & \cellcolor[HTML]{D7E082}64.89\% & \cellcolor[HTML]{FEEA83}60.80\% & \cellcolor[HTML]{FDD37F}55.37\% & \cellcolor[HTML]{FBA276}44.23\% \\
\textbf{vitl32}  & \cellcolor[HTML]{63BE7B}76.46\% & \cellcolor[HTML]{70C27C}75.25\% & \cellcolor[HTML]{7AC57D}74.17\% & \cellcolor[HTML]{87C97E}72.90\% & \cellcolor[HTML]{93CC7E}71.68\% & \cellcolor[HTML]{A4D17F}70.06\% & \cellcolor[HTML]{B9D780}67.90\% & \cellcolor[HTML]{D7E082}64.93\% & \cellcolor[HTML]{FEE783}60.06\% & \cellcolor[HTML]{FCB97A}49.47\% \\
\textbf{vitb16}  & \cellcolor[HTML]{9CCF7F}70.78\% & \cellcolor[HTML]{A8D27F}69.66\% & \cellcolor[HTML]{AFD480}68.90\% & \cellcolor[HTML]{C0D981}67.17\% & \cellcolor[HTML]{C9DC81}66.35\% & \cellcolor[HTML]{DCE182}64.39\% & \cellcolor[HTML]{F1E784}62.32\% & \cellcolor[HTML]{FEE582}59.57\% & \cellcolor[HTML]{FDD07E}54.79\% & \cellcolor[HTML]{FA9F75}43.46\% \\
\textbf{vitb32}  & \cellcolor[HTML]{F2E884}62.18\% & \cellcolor[HTML]{F9EA84}61.53\% & \cellcolor[HTML]{FFEB84}60.93\% & \cellcolor[HTML]{FEE583}59.69\% & \cellcolor[HTML]{FEDE81}57.98\% & \cellcolor[HTML]{FDD780}56.46\% & \cellcolor[HTML]{FDCB7D}53.56\% & \cellcolor[HTML]{FCB97A}49.53\% & \cellcolor[HTML]{FA9F75}43.61\% & \cellcolor[HTML]{F8696B}30.98\% \\
\hline
\hline
\textbf{swinV2b} & \cellcolor[HTML]{CEDD82}65.85\% & \cellcolor[HTML]{CADC81}66.18\% & \cellcolor[HTML]{D7E082}64.88\% & \cellcolor[HTML]{E0E283}64.01\% & \cellcolor[HTML]{E9E583}63.11\% & \cellcolor[HTML]{F9EA84}61.48\% & \cellcolor[HTML]{FEE683}59.87\% & \cellcolor[HTML]{FEDC81}57.44\% & \cellcolor[HTML]{FDCA7D}53.46\% & \cellcolor[HTML]{FBA476}44.59\% \\
\textbf{swinV2s} & \cellcolor[HTML]{DEE283}64.18\% & \cellcolor[HTML]{E6E483}63.46\% & \cellcolor[HTML]{ECE683}62.77\% & \cellcolor[HTML]{F7E984}61.69\% & \cellcolor[HTML]{FEE983}60.50\% & \cellcolor[HTML]{FEE382}59.21\% & \cellcolor[HTML]{FEDD81}57.79\% & \cellcolor[HTML]{FDD17F}55.02\% & \cellcolor[HTML]{FCC17B}51.26\% & \cellcolor[HTML]{FA9473}40.91\% \\
\textbf{swinV2t} & \cellcolor[HTML]{FEE683}59.78\% & \cellcolor[HTML]{FEE382}59.05\% & \cellcolor[HTML]{FEDE81}58.07\% & \cellcolor[HTML]{FDD680}56.26\% & \cellcolor[HTML]{FDD17F}54.99\% & \cellcolor[HTML]{FDCB7D}53.56\% & \cellcolor[HTML]{FCC27C}51.62\% & \cellcolor[HTML]{FCB479}48.30\% & \cellcolor[HTML]{FAA075}43.69\% & \cellcolor[HTML]{F8756D}33.88\% \\
\hline
\end{tabular}
\label{tab:top1_accuracy_cifar100}
\end{table}


\begin{table}[ht]
\centering
\footnotesize
\caption{Accuracy difference (\%) of the active learning strategies explained in Secion \ref{subsec:Active_Learning_Meth} relative to the \textit{random} strategy on CIFAR10 under varying label noise rates (0.0--0.9) on CIFAR10 across all models at 13\% labeled data proportion. Positive values (green) indicate improvement over \textit{random}, while negative values (red) indicate lower accuracy. GCI\_ViTAL generally outperforms the other strategies, except at extreme noise levels}
\label{tab:acc_diff_cifar10}
\begin{tabular}{|l|*{10}{c|}}
\hline
\textbf{Strategy} & \multicolumn{10}{c|}{\textbf{Label Noise Rate}} \\
\cline{2-11}
                  & \textbf{0.0} 
                  & \textbf{0.1} 
                  & \textbf{0.2} 
                  & \textbf{0.3} 
                  & \textbf{0.4} 
                  & \textbf{0.5} 
                  & \textbf{0.6} 
                  & \textbf{0.7} 
                  & \textbf{0.8} 
                  & \textbf{0.9} \\
\hline
\textbf{random}   
& \cellcolor[HTML]{FCFCFF}0.00\% 
& \cellcolor[HTML]{FCFCFF}0.00\% 
& \cellcolor[HTML]{FCFCFF}0.00\% 
& \cellcolor[HTML]{FCFCFF}0.00\% 
& \cellcolor[HTML]{FCFCFF}0.00\% 
& \cellcolor[HTML]{FCFCFF}0.00\% 
& \cellcolor[HTML]{FCFCFF}0.00\% 
& \cellcolor[HTML]{FCFCFF}0.00\% 
& \cellcolor[HTML]{FCFCFF}0.00\% 
& \cellcolor[HTML]{FCFCFF}0.00\% \\
\hline
\textbf{GCI\_ViTAL} 
& \cellcolor[HTML]{E3F2EA}0.19\% 
& \cellcolor[HTML]{E3F2EA}0.19\% 
& \cellcolor[HTML]{CAE8D4}0.38\% 
& \cellcolor[HTML]{E8F4EE}0.16\% 
& \cellcolor[HTML]{D1EBDA}0.33\% 
& \cellcolor[HTML]{F99597}-0.32\% 
& \cellcolor[HTML]{F0F7F5}0.10\% 
& \cellcolor[HTML]{EFF7F4}0.10\% 
& \cellcolor[HTML]{FAD5D8}-0.12\% 
& \cellcolor[HTML]{A5D9B4}0.67\% \\
\hline
\textbf{entropy}  
& \cellcolor[HTML]{FACBCE}-0.15\% 
& \cellcolor[HTML]{F8696B}-0.45\% 
& \cellcolor[HTML]{E0F1E7}0.22\% 
& \cellcolor[HTML]{FACACD}-0.15\% 
& \cellcolor[HTML]{F87678}-0.41\% 
& \cellcolor[HTML]{F99EA1}-0.29\% 
& \cellcolor[HTML]{FBDBDD}-0.10\% 
& \cellcolor[HTML]{DCEFE4}0.25\% 
& \cellcolor[HTML]{E3F2EA}0.19\% 
& \cellcolor[HTML]{63BE7B}1.16\% \\
\hline
\end{tabular}
\end{table}


\begin{table}[ht]
\centering
\footnotesize
\caption{This table shows accuracy differences (\%) of the active learning strategies explained in Secion \ref{subsec:Active_Learning_Meth}, relative to \textit{random} query, under varying label noise rates (0.0--0.9), at 13\% labeled data proportion on CIFAR100, across all models. Positive values indicate improvement, while negative values indicate lower accuracy than \textit{random}. We see that \textit{GCI\_ViTAL} dominates for moderate noise rates (0.1--0.6), and both \textit{GCI\_ViTAL} and \textit{entropy} show higher accuracy than \textit{random} in the mid-range of noise levels.}
\label{tab:acc_diff_random}
\begin{tabular}{|l|cccccccccc|}
\hline
\textbf{Strategy} & \multicolumn{10}{c|}{\textbf{Label Noise Rate}} \\
\cline{2-11}
                  & \textbf{0.0} 
                  & \textbf{0.1} 
                  & \textbf{0.2} 
                  & \textbf{0.3} 
                  & \textbf{0.4} 
                  & \textbf{0.5} 
                  & \textbf{0.6} 
                  & \textbf{0.7} 
                  & \textbf{0.8} 
                  & \textbf{0.9} \\
\hline
\textbf{random}      
& \cellcolor[HTML]{FCFCFF}0.00\% 
& \cellcolor[HTML]{FCFCFF}0.00\% 
& \cellcolor[HTML]{FCFCFF}0.00\% 
& \cellcolor[HTML]{FCFCFF}0.00\% 
& \cellcolor[HTML]{FCFCFF}0.00\% 
& \cellcolor[HTML]{FCFCFF}0.00\% 
& \cellcolor[HTML]{FCFCFF}0.00\% 
& \cellcolor[HTML]{FCFCFF}0.00\% 
& \cellcolor[HTML]{FCFCFF}0.00\% 
& \cellcolor[HTML]{FCFCFF}0.00\% \\

\textbf{GCI\_ViTAL}
& \cellcolor[HTML]{FACBCE}-0.15\% 
& \cellcolor[HTML]{C4E6CF}0.36\% 
& \cellcolor[HTML]{C9E7D3}0.33\% 
& \cellcolor[HTML]{C8E7D2}0.34\% 
& \cellcolor[HTML]{D5ECDD}0.25\% 
& \cellcolor[HTML]{B1DEBE}0.48\% 
& \cellcolor[HTML]{F1F8F5}0.07\% 
& \cellcolor[HTML]{F8696B}-0.46\% 
& \cellcolor[HTML]{F9A9AC}-0.26\% 
& \cellcolor[HTML]{63BE7B}0.98\% \\

\textbf{entropy}     
& \cellcolor[HTML]{F98E90}-0.34\% 
& \cellcolor[HTML]{F99799}-0.31\% 
& \cellcolor[HTML]{E7F4ED}0.13\% 
& \cellcolor[HTML]{C2E5CD}0.37\% 
& \cellcolor[HTML]{E7F4ED}0.14\% 
& \cellcolor[HTML]{FBFAFD}0.00\% 
& \cellcolor[HTML]{A2D8B2}0.57\% 
& \cellcolor[HTML]{F0F7F5}0.08\% 
& \cellcolor[HTML]{FAC3C5}-0.18\% 
& \cellcolor[HTML]{BEE3CA}0.40\% \\
\hline
\end{tabular}
\end{table}

\begin{figure}[ht!]
    \centering
    \includegraphics[width=\textwidth]{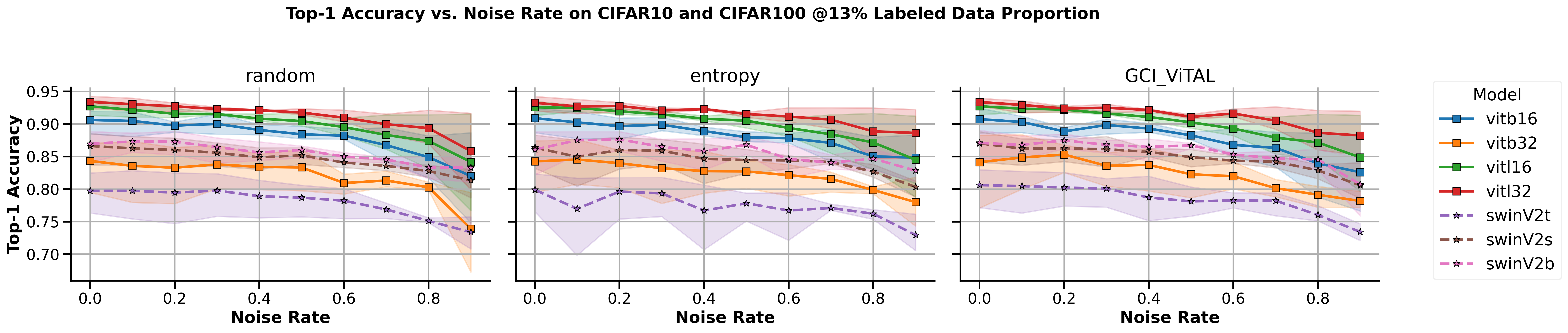} 
    \caption{Top-1 Accuracy vs. Label Noise Rate averaged over CIFAR10 and CIFAR100 at 13\% labeled data. Each subplot shows a different active learning strategy (random, entropy, GCI\_ViTAL) as explained in Section \ref{subsec:Active_Learning_Meth}, across various Vision Transformer (ViT) and Swin Transformer (SwinV2) models. See Appendix \ref{sec:appendix_a_results} for the version of this graph split by dataset and at different labeled data proportions.}
    \label{fig:final_accuracy_vs_noise_rate_cifar100_13}
\end{figure}

\begin{itemize}
    \item In Tables \ref{tab:top1_accuracy_cifar10} and \ref{tab:top1_accuracy_cifar100} corresponding to CIFAR10 and CIFAR100 results, we observe that top-1 accuracy declines with increasing label noise across all models. Larger models (ViTl and SwinV2b) maintain relatively higher accuracy than smaller models (ViTb and SwinV2t), even at high label noise rates. Notably, ViTl32 achieves 64.93\% top-1 accuracy at a 70\% noise level, surpassing ViTb32 (62.18\%) even when ViTb32 is trained with only clean labels. We see the same difference between the  SwinV2t and SwinV2b models at 60\% and 0\% label noise.

    \item Interestingly, we see that ViTl32 consistently outperforms the ViTl16 variant, even though smaller patch sizes are expected to allow for more detailed image comprehension and superior results. This is unexpected because the base models with 16×16 and 32×32 patch sizes adhere to the standard performance order.

    \item In Tables \ref{tab:acc_diff_cifar10} and \ref{tab:acc_diff_random} we observe that the difference in performance between the three DAL strategies is most pronounced at mid-range noise rates (30-60\%). In these ranges, GCI\_ViTAL shows more robustness to label noise relatively on both CIFAR10 and CIFAR100. On CIFAR100, both entropy and GCI\_ViTAL outperform the random baseline in the moderate noise rate ranges (30-60\%). At very high noise rates, there is no advantage in smarter sampling as there are just very few clean labels irrespective of how informative the samples are themselves. We expand these results in Figure \ref{fig:final_accuracy_vs_noise_rate_cifar100_13}, splitting the performance by ViT variant.
\end{itemize}

\subsection{Model Calibration}
We measure model calibration using the Brier Score, as explained in section \ref{subsec:Evaluation}. For the Brier score, unlike accuracy, the lower the Brier score the better calibrated the model is.

\begin{table}[ht]
\centering
\footnotesize
\caption{
Brier Score (\%), given by Equation \ref{eq:brier_score} for various Vision Transformer (ViT) and SwinV2 models under increasing label noise rates on CIFAR10, averaged over all Active Learning strategies and labeled data proportions. Lower scores indicate better model calibration. As expected, calibration degrades with increasing label noise, and larger models are better calibrated than smaller models. The impact of patch size on model calibration remains non-trivial since the ViT base and large variants show inconsistent results on varying patch sizes.  
}

\begin{tabular}{|l|*{10}{c|}}
\hline
\textbf{Model} & \multicolumn{10}{c|}{\textbf{Label Noise Rate}} \\
\cline{2-11}
               & \textbf{0} & \textbf{0.1} & \textbf{0.2} & \textbf{0.3} & \textbf{0.4} & \textbf{0.5} & \textbf{0.6} & \textbf{0.7} & \textbf{0.8} & \textbf{0.9} \\
\hline
\textbf{vitl16}  
& \cellcolor[HTML]{6AC07B}10.38\% 
& \cellcolor[HTML]{79C47C}12.74\% 
& \cellcolor[HTML]{9ACE7E}18.19\% 
& \cellcolor[HTML]{A8D27F}20.49\% 
& \cellcolor[HTML]{D6DF81}27.88\% 
& \cellcolor[HTML]{FFE984}35.17\% 
& \cellcolor[HTML]{FED380}41.32\% 
& \cellcolor[HTML]{FCB179}51.07\% 
& \cellcolor[HTML]{FB9874}58.02\% 
& \cellcolor[HTML]{F9776E}67.44\% \\
\textbf{vitl32}  
& \cellcolor[HTML]{63BE7B}9.10\% 
& \cellcolor[HTML]{75C37C}12.18\% 
& \cellcolor[HTML]{95CC7D}17.41\% 
& \cellcolor[HTML]{A4D07E}19.72\% 
& \cellcolor[HTML]{D5DF81}27.82\% 
& \cellcolor[HTML]{F3E783}32.70\% 
& \cellcolor[HTML]{FFDD82}38.69\% 
& \cellcolor[HTML]{FDBB7B}48.28\% 
& \cellcolor[HTML]{FB9F76}55.93\% 
& \cellcolor[HTML]{FA8070}64.88\% \\
\textbf{vitb16}  
& \cellcolor[HTML]{78C47C}12.65\% 
& \cellcolor[HTML]{94CC7D}17.13\% 
& \cellcolor[HTML]{B5D57F}22.58\% 
& \cellcolor[HTML]{C9DB80}25.77\% 
& \cellcolor[HTML]{FBE983}33.86\% 
& \cellcolor[HTML]{FED981}39.77\% 
& \cellcolor[HTML]{FDC67D}45.08\% 
& \cellcolor[HTML]{FCA978}53.20\% 
& \cellcolor[HTML]{FA8D72}61.18\% 
& \cellcolor[HTML]{F9746E}68.08\% \\
\textbf{vitb32}  
& \cellcolor[HTML]{99CD7E}17.92\% 
& \cellcolor[HTML]{A7D17E}20.24\% 
& \cellcolor[HTML]{C1D980}24.49\% 
& \cellcolor[HTML]{D2DE81}27.28\% 
& \cellcolor[HTML]{F7E883}33.23\% 
& \cellcolor[HTML]{FFDB81}39.16\% 
& \cellcolor[HTML]{FDC17C}46.53\% 
& \cellcolor[HTML]{FCAD78}52.14\% 
& \cellcolor[HTML]{FB9273}59.60\% 
& \cellcolor[HTML]{F9716D}68.90\% \\
\hline\hline
\textbf{swinV2b} 
& \cellcolor[HTML]{8BC97D}15.62\% 
& \cellcolor[HTML]{92CB7D}16.89\% 
& \cellcolor[HTML]{ADD37F}21.28\% 
& \cellcolor[HTML]{C8DB80}25.66\% 
& \cellcolor[HTML]{E0E282}29.53\% 
& \cellcolor[HTML]{FFE684}35.96\% 
& \cellcolor[HTML]{FECF7F}42.64\% 
& \cellcolor[HTML]{FDB47A}50.03\% 
& \cellcolor[HTML]{FB9874}58.16\% 
& \cellcolor[HTML]{F9726D}68.69\% \\
\textbf{swinV2s} 
& \cellcolor[HTML]{91CB7D}16.75\% 
& \cellcolor[HTML]{A2D07E}19.40\% 
& \cellcolor[HTML]{B7D67F}22.83\% 
& \cellcolor[HTML]{D2DE81}27.21\% 
& \cellcolor[HTML]{F0E683}32.15\% 
& \cellcolor[HTML]{FFDF82}38.00\% 
& \cellcolor[HTML]{FEC87E}44.42\% 
& \cellcolor[HTML]{FCB179}51.08\% 
& \cellcolor[HTML]{FB8F73}60.43\% 
& \cellcolor[HTML]{F9726D}68.81\% \\
\textbf{swinV2t} 
& \cellcolor[HTML]{B4D57F}22.44\% 
& \cellcolor[HTML]{C0D880}24.29\% 
& \cellcolor[HTML]{CFDD81}26.72\% 
& \cellcolor[HTML]{E7E482}30.67\% 
& \cellcolor[HTML]{FFE884}35.45\% 
& \cellcolor[HTML]{FED580}40.89\% 
& \cellcolor[HTML]{FDBE7C}47.23\% 
& \cellcolor[HTML]{FCAA78}53.04\% 
& \cellcolor[HTML]{FA8B72}61.64\% 
& \cellcolor[HTML]{F8696B}71.14\% \\
\hline
\end{tabular}
\label{tab:brier_score_cifar10}
\end{table}

\begin{table}[ht]
\centering
\footnotesize
\caption{
Brier Score (\%) for various Vision Transformer (ViT) and SwinV2 models under increasing label noise rates on CIFAR100, averaged over all active learning strategies and labeled data proportion experiments. Lower Brier scores indicate better model calibration. As expected, calibration degrades with increasing label noise. We see that ViTl32 consistently maintains better calibration compared to other ViT models, suggesting better uncertainty estimation under label noise. Large Vits over large SwinV2 transformers when it comes to calibration.
}

\begin{tabular}{|l|*{10}{c|}}
\hline
\textbf{Model} & \multicolumn{10}{c|}{\textbf{Label Noise Rate}} \\
\cline{2-11}
               & \textbf{0} & \textbf{0.1} & \textbf{0.2} & \textbf{0.3} & \textbf{0.4} & \textbf{0.5} & \textbf{0.6} & \textbf{0.7} & \textbf{0.8} & \textbf{0.9} \\
\hline
\textbf{vitl16} &
\cellcolor[HTML]{67BF7B}34.22\% & \cellcolor[HTML]{7AC47C}37.56\% & \cellcolor[HTML]{91CB7D}41.73\% & 
\cellcolor[HTML]{A9D27F}46.15\% & \cellcolor[HTML]{C6DA80}51.45\% & \cellcolor[HTML]{E3E282}56.61\% & 
\cellcolor[HTML]{FFE884}62.54\% & \cellcolor[HTML]{FEC87E}69.89\% & \cellcolor[HTML]{FCAA78}76.92\% & 
\cellcolor[HTML]{FA8771}85.22\% \\

\textbf{vitl32} &
\cellcolor[HTML]{63BE7B}33.34\% & \cellcolor[HTML]{71C27B}35.94\% & \cellcolor[HTML]{84C77C}39.47\% &
\cellcolor[HTML]{9BCE7E}43.56\% & \cellcolor[HTML]{B5D57F}48.29\% & \cellcolor[HTML]{D0DD81}53.27\% &
\cellcolor[HTML]{F3E783}59.58\% & \cellcolor[HTML]{FED680}66.58\% & \cellcolor[HTML]{FDB57A}74.54\% &
\cellcolor[HTML]{FB8F73}83.35\% \\

\textbf{vitb16} &
\cellcolor[HTML]{8BC97D}40.64\% & \cellcolor[HTML]{99CD7E}43.26\% & \cellcolor[HTML]{ADD37F}46.79\% &
\cellcolor[HTML]{C7DA80}51.50\% & \cellcolor[HTML]{DDE182}55.60\% & \cellcolor[HTML]{FBE983}60.94\% &
\cellcolor[HTML]{FFDA81}65.87\% & \cellcolor[HTML]{FDBF7C}72.14\% & \cellcolor[HTML]{FBA176}79.12\% &
\cellcolor[HTML]{FA7E70}87.24\% \\

\textbf{vitb32} &
\cellcolor[HTML]{C5DA80}51.24\% & \cellcolor[HTML]{CDDC81}52.70\% & \cellcolor[HTML]{D9E081}54.91\% &
\cellcolor[HTML]{ECE582}58.28\% & \cellcolor[HTML]{FFE884}62.40\% & \cellcolor[HTML]{FED781}66.50\% &
\cellcolor[HTML]{FDBE7C}72.27\% & \cellcolor[HTML]{FCA377}78.70\% & \cellcolor[HTML]{FA8671}85.43\% &
\cellcolor[HTML]{F8696B}92.15\% \\
\hline
\hline
\textbf{swinV2b} &
\cellcolor[HTML]{AED37F}47.08\% & \cellcolor[HTML]{B4D57F}48.14\% & \cellcolor[HTML]{C2D980}50.67\% &
\cellcolor[HTML]{D4DE81}54.02\% & \cellcolor[HTML]{F6E883}60.18\% & \cellcolor[HTML]{FFE683}62.99\% &
\cellcolor[HTML]{FECF7F}68.39\% & \cellcolor[HTML]{FDB47A}74.64\% & \cellcolor[HTML]{FB9474}82.10\% &
\cellcolor[HTML]{F9756E}89.56\% \\

\textbf{swinV2s} &
\cellcolor[HTML]{BED880}49.94\% & \cellcolor[HTML]{C5DA80}51.17\% & \cellcolor[HTML]{D1DD81}53.38\% &
\cellcolor[HTML]{E8E482}57.60\% & \cellcolor[HTML]{FFEB84}61.70\% & \cellcolor[HTML]{FED781}66.42\% &
\cellcolor[HTML]{FDBA7B}73.24\% & \cellcolor[HTML]{FCA477}78.52\% & \cellcolor[HTML]{FA8C72}84.10\% &
\cellcolor[HTML]{F96E6C}91.09\% \\

\textbf{swinV2t} &
\cellcolor[HTML]{D4DE81}54.00\% & \cellcolor[HTML]{DDE182}55.53\% & \cellcolor[HTML]{EFE683}58.81\% &
\cellcolor[HTML]{FEEA83}61.60\% & \cellcolor[HTML]{FFDA81}65.79\% & \cellcolor[HTML]{FEC87E}70.00\% &
\cellcolor[HTML]{FDB67A}74.10\% & \cellcolor[HTML]{FB9C75}80.22\% & \cellcolor[HTML]{FA8370}86.16\% &
\cellcolor[HTML]{F96D6C}91.28\% \\
\bottomrule
\end{tabular}
\label{tab:brier_score_cifar100}
\end{table}


\begin{table}[ht]
\centering
\footnotesize
\caption{Brier score differences (\%) of active learning strategies relative to \textit{random} selection under varying label noise rates (0.0--0.9) at 13\% labeled data proportion on CIFAR10. In the table, negative values mean the strategy in that row has higher brier scores than the random query strategy since we calculate the difference as ($str_{random} - str_K$), where K is one of the three strategies in Section \ref{subsec:Active_Learning_Meth}. Green represents good calibration, and red represents bad calibration. The \textit{entropy} strategy provides slight improvements at higher noise levels, while \textit{GCI\_ViTAL} struggles with significantly worse calibration, particularly as noise increases.}
\label{tab:brier_cifar10}
\begin{tabular}{|l|cccccccccc|}
\hline
\textbf{Strategy} & \multicolumn{10}{c|}{\textbf{Label Noise Rate}} \\
\cline{2-11}
                  & \textbf{0.0} 
                  & \textbf{0.1} 
                  & \textbf{0.2} 
                  & \textbf{0.3} 
                  & \textbf{0.4} 
                  & \textbf{0.5} 
                  & \textbf{0.6} 
                  & \textbf{0.7} 
                  & \textbf{0.8} 
                  & \textbf{0.9} \\
\hline
\textbf{random}      
& \cellcolor[HTML]{74C58A}0.00\% 
& \cellcolor[HTML]{74C58A}0.00\% 
& \cellcolor[HTML]{74C58A}0.00\% 
& \cellcolor[HTML]{74C58A}0.00\% 
& \cellcolor[HTML]{74C58A}0.00\% 
& \cellcolor[HTML]{74C58A}0.00\% 
& \cellcolor[HTML]{74C58A}0.00\% 
& \cellcolor[HTML]{74C58A}0.00\% 
& \cellcolor[HTML]{74C58A}0.00\% 
& \cellcolor[HTML]{74C58A}0.00\% \\

\textbf{GCI\_ViTAL}
& \cellcolor[HTML]{FAD3D6}-1.34\% 
& \cellcolor[HTML]{FAC2C5}-1.63\% 
& \cellcolor[HTML]{F9A8AB}-2.06\% 
& \cellcolor[HTML]{FBDBDD}-1.21\% 
& \cellcolor[HTML]{F99598}-2.38\% 
& \cellcolor[HTML]{F8696B}-3.14\% 
& \cellcolor[HTML]{F99EA0}-2.24\% 
& \cellcolor[HTML]{FAB3B6}-1.87\% 
& \cellcolor[HTML]{F9A1A3}-2.18\% 
& \cellcolor[HTML]{FBECEF}-0.92\% \\

\textbf{entropy}     
& \cellcolor[HTML]{FBF8FB}-0.72\% 
& \cellcolor[HTML]{FBEBEE}-0.93\% 
& \cellcolor[HTML]{B9E1C5}-0.33\% 
& \cellcolor[HTML]{EFF7F4}-0.59\% 
& \cellcolor[HTML]{FACBCE}-1.47\% 
& \cellcolor[HTML]{A4D9B4}-0.23\% 
& \cellcolor[HTML]{FBF7FA}-0.73\% 
& \cellcolor[HTML]{ABDBB9}-0.26\% 
& \cellcolor[HTML]{FBE3E6}-1.07\% 
& \cellcolor[HTML]{63BE7B}0.08\% \\
\hline
\end{tabular}
\end{table}


\begin{table}[ht]
\centering
\footnotesize
\caption{Calibration Score differences (\%) of the active learning strategies explained in Section \ref{subsec:Active_Learning_Meth} relative to \textit{random} selection under varying label noise rates (0.0--0.9) at 13\% labeled data proportion on CIFAR100. Lower values (green) indicate better calibration. The \textit{entropy}-based strategy generally improves calibration under moderate to high noise levels, while \textit{GCI\_ViTAL} struggles under high noise rates.}
\label{tab:calibration_diff_cifar100}
\begin{tabular}{|l|cccccccccc|}
\hline
\textbf{Strategy} & \multicolumn{10}{c|}{\textbf{Label Noise Rate}} \\
\cline{2-11}
                  & \textbf{0.0} 
                  & \textbf{0.1} 
                  & \textbf{0.2} 
                  & \textbf{0.3} 
                  & \textbf{0.4} 
                  & \textbf{0.5} 
                  & \textbf{0.6} 
                  & \textbf{0.7} 
                  & \textbf{0.8} 
                  & \textbf{0.9} \\
\hline
\textbf{random}      
& \cellcolor[HTML]{FCFCFF}0.00\% 
& \cellcolor[HTML]{FCFCFF}0.00\% 
& \cellcolor[HTML]{FCFCFF}0.00\% 
& \cellcolor[HTML]{FCFCFF}0.00\% 
& \cellcolor[HTML]{FCFCFF}0.00\% 
& \cellcolor[HTML]{FCFCFF}0.00\% 
& \cellcolor[HTML]{FCFCFF}0.00\% 
& \cellcolor[HTML]{FCFCFF}0.00\% 
& \cellcolor[HTML]{FCFCFF}0.00\% 
& \cellcolor[HTML]{FCFCFF}0.00\% \\

\textbf{GCI\_ViTAL}
& \cellcolor[HTML]{FACBCE}-0.60\% 
& \cellcolor[HTML]{FBEEF1}-0.24\% 
& \cellcolor[HTML]{FAD2D4}-0.56\% 
& \cellcolor[HTML]{FAC7CA}-0.67\% 
& \cellcolor[HTML]{FAB9BC}-0.83\% 
& \cellcolor[HTML]{FBEFF2}-0.22\% 
& \cellcolor[HTML]{FAB4B6}-0.90\% 
& \cellcolor[HTML]{F8696B}-1.75\% 
& \cellcolor[HTML]{F99597}-1.24\% 
& \cellcolor[HTML]{FACACC}-0.65\% \\

\textbf{entropy}     
& \cellcolor[HTML]{FBDEE1}-0.42\% 
& \cellcolor[HTML]{FBE0E3}-0.40\% 
& \cellcolor[HTML]{D2EBDB}0.07\% 
& \cellcolor[HTML]{63BE7B}0.46\% 
& \cellcolor[HTML]{F9FBFC}-0.07\% 
& \cellcolor[HTML]{85CC98}0.34\% 
& \cellcolor[HTML]{86CD99}0.33\% 
& \cellcolor[HTML]{FBF7FA}-0.13\% 
& \cellcolor[HTML]{FBFAFD}-0.10\% 
& \cellcolor[HTML]{FBE3E6}-0.36\% \\
\hline
\end{tabular}
\end{table}

\begin{figure}[ht!]
    \centering
    \includegraphics[width=\textwidth]{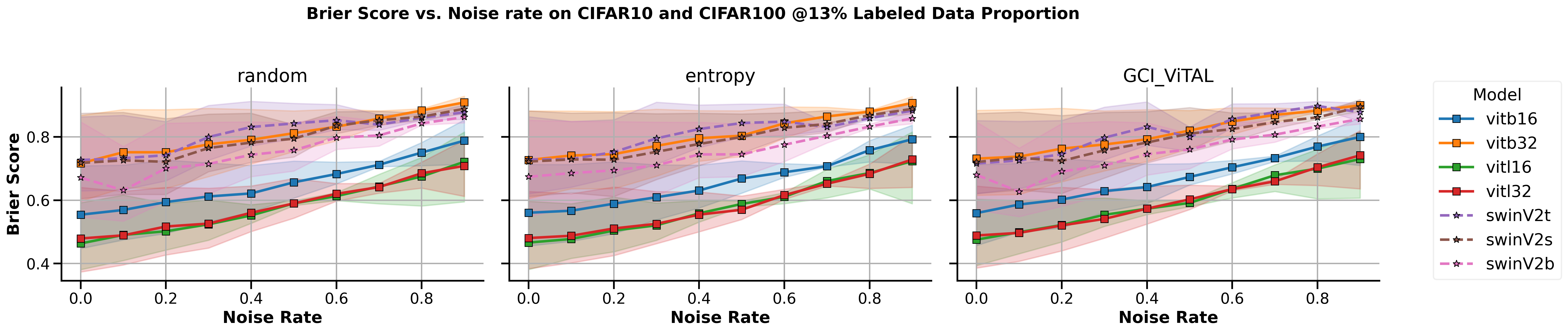} 
    \caption{The Brier Score vs. Noise Rate averaged over both CIFAR10 and CIFAR100 at 13\% labeled data proportion. Each subplot compares multiple ViT and SwinV2 models under different Active learning strategies. We see an overall dominance of the ViT architecture over the SwinV2 variants. We also see marginally higher calibration using the random query over the information-based acquisition strategies. See appendix \ref{sec:appendix_a_results} for similar additional results}
    \label{fig:final_brier_vs_noise_rate_cifar100_13}
\end{figure}

\begin{itemize}
    \item In tables \ref{tab:brier_score_cifar10} and \ref{tab:brier_score_cifar100} corresponding to CIFAR10 and CIFAR100 Brier scores, we observe that as expected, the Brier score worsens with increasing label noise. Despite losing overall accuracy under heavy noise, some models manage to stay relatively more calibrated than others (e.g., ViTl32 vs. ViTl16).

    \item Tables \ref{tab:brier_cifar10} and \ref{tab:calibration_diff_cifar100} show that random selection maintains better calibration than GCI\_ViTAL across most noise levels, especially as label noise increases, while entropy-based selection occasionally outperforms random at high noise rates but remains inconsistent in the mid-range. This comes as a surprise result since it is common to expect high accuracy to lead to high model calibration. However, we think this can be explained by the fact that most DAL strategies are developed to optimize accuracy and not calibration.

    \item In general, the ViT variants except for ViTb32, maintain better calibration than the SwinV2 transformer models across noise rates and DAL acquisition strategies as shown in Figure \ref{fig:final_brier_vs_noise_rate_cifar100_13}. The calibration curves are averaged over CIFAR10 and CIFAR100 at 13\% labeled data proportion. We include additional Brier score curves split by dataset at different data proportions in Appendix \ref{sec:appendix_a_results}.

\end{itemize}

\subsection{Model Efficiency}\label{subsec:efficiency_results}

\begin{itemize}
    \item As expected, training times grow almost linearly with the amount of labeled data across all models irrespective of the DAL strategy or label noise rate. The $16 \times 16$ patch size models have notably higher training times than the $32 \times 32$ variant in both the base and large ViTs as expected since smaller patch sizes result in more tokens, forcing the transformer’s self-attention mechanism to handle more token interactions, leading to higher computational costs. We include the average training times per model in table \ref{table:models} and show the graph over the percentage of labeled data in Appendix \ref{sec:appendix_a_results}.

    \item We note the clear inefficiency of the ViTl16 as compared to ViTl32 is not justified in both accuracy and calibration, meaning all else being equal one should use the ViTl32. We see a similar trend between ViTs and SwinV2 transformers in that they post comparable training times but the SwinV2 transformers lag in both accuracy and calibration as shown in Tables \ref{tab:top1_accuracy_cifar10},\ref{tab:top1_accuracy_cifar100} and \ref{tab:brier_score_cifar10}, \ref{tab:brier_score_cifar100} respectively, under label noise across all DAL strategies shown in Figure \ref{fig:final_accuracy_vs_noise_rate_cifar100_13}.
\end{itemize}

\section{Conclusion}\label{sec:Conclusion}
In conclusion, practical considerations such as resource constraints and the labeling budget play a crucial role in choosing Vision Transformer (ViT) and Swin Transformer architectures for learning under label noise. To this end, this study experimentally investigated how different Vision Transformer and Swin Transformer configurations in patch and embedding sizes perform under varying label noise rates and labeling budgets. A key finding is that larger ViTs, even with bigger patch sizes (ViTl32) generally perform better against moderate label noise, providing higher accuracy and stronger calibration than smaller ViTs or SwinV2 models. Surprisingly, despite having a smaller patch size, the inefficient ViTl16 offers no significant advantage in accuracy or calibration while incurring higher computational costs than ViTl32. We find that SwinV2 transformers train at comparable speeds to the ViTs, but lag in calibration and accuracy under label noise. We find that information-based DAL strategies like entropy and GCI\_ViTAL while leading to more accurate models at low data proportions and moderate label noise rates, offer little to no performance gains at very high noise rates, where the data is too corrupted. The random selection strategy often maintains better calibration than both entropy and GCI\_ViTAL, especially as noise levels increase, highlighting the limitations of DAL strategies that primarily optimize for accuracy rather than uncertainty calibration. Overall, limited to our experimental setting, ViTl32 emerges as the most efficient and robust ViT variant in balancing accuracy, calibration, and computational efficiency. In future work, we plan to explore the application of ViTs in autonomous driving.

\section*{Acknowledgments}
We acknowledge technical support and advanced computing resources from University of Hawaii Information Technology Services – Research Cyberinfrastructure, funded in part by the National Science Foundation CC* awards \texttt{\#}2201428 and \texttt{\#}2232862.

\bibliographystyle{unsrt}  
\bibliography{main}  

\clearpage
\appendix
\section*{Appendix}\label{sec:appendix_a_results}
We include additional plots for accuracy, calibration, and label noise. We vary the data proportion used in most of the diagrams.

\subsection*{CIFAR10 Variable Label Noise, Accuracy, and Calibration Curves}

\begin{figure}[htbp]
    \centering
    \includegraphics[width=\textwidth]{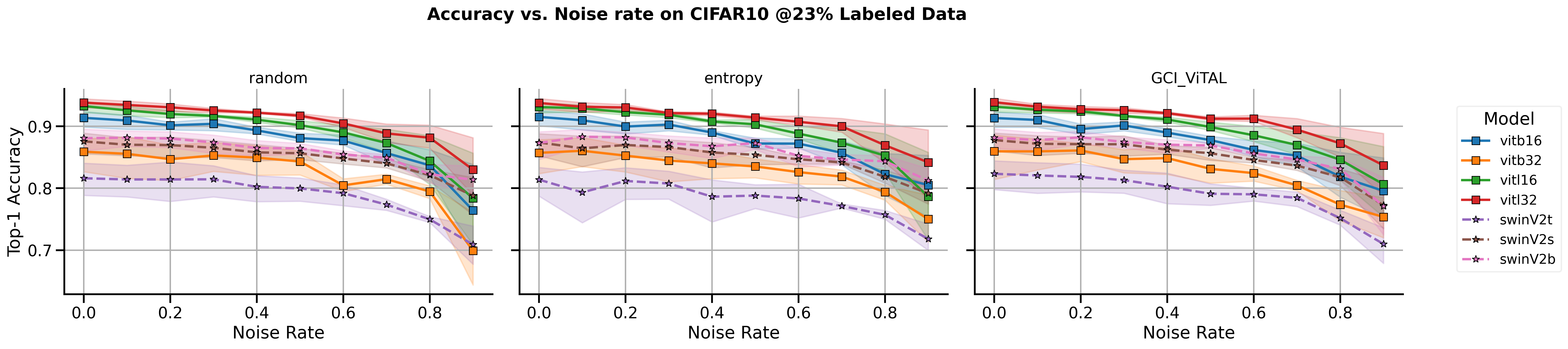}
    \caption{Top-1 Accuracy vs. Noise Rate on CIFAR10 with 23\% labeled data. Each subplot represents a different DAL strategy (random, entropy, GCI\_ViTAL) applied to Vision Transformer (ViT) and Swin Transformer (Swin) models. The same trends seen at 13\% labeled data persist across VIT size and AL strategy.}
\end{figure}

\begin{figure}[htbp]
    \centering
    \includegraphics[width=\textwidth]{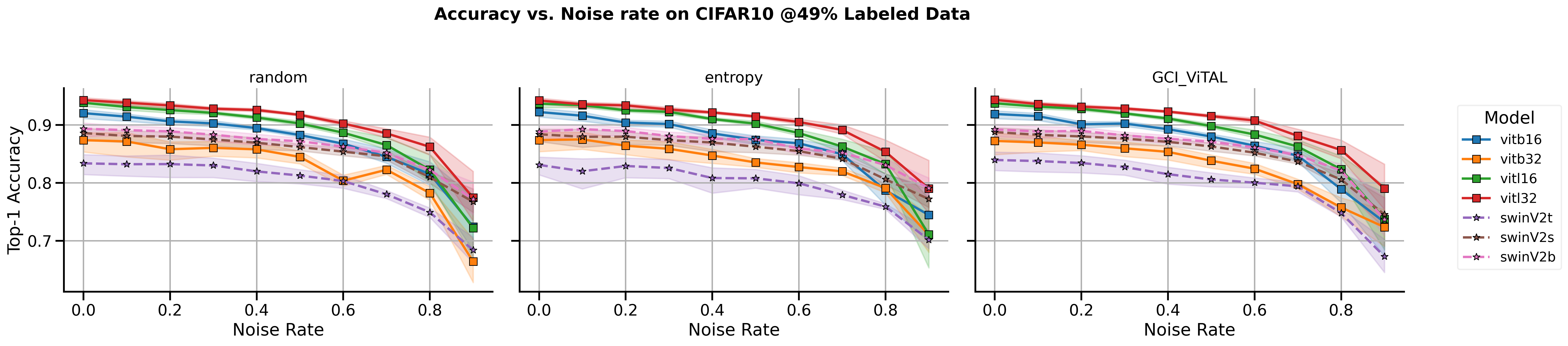}
    \caption{Top-1 Accuracy vs. Noise Rate on CIFAR10 with 49\% labeled data. Increasing the labeled data proportion improves accuracy across all models, but the trends remain consistent with the 13\% and 23\% label proportion setting: larger ViTs over smaller variants, ViTs over SwinV2 models.}
\end{figure}

\begin{figure}[htbp]\label{fig;final_brier_vs_noise_rate_cifar10_23}
    \centering
    \includegraphics[width=\textwidth]{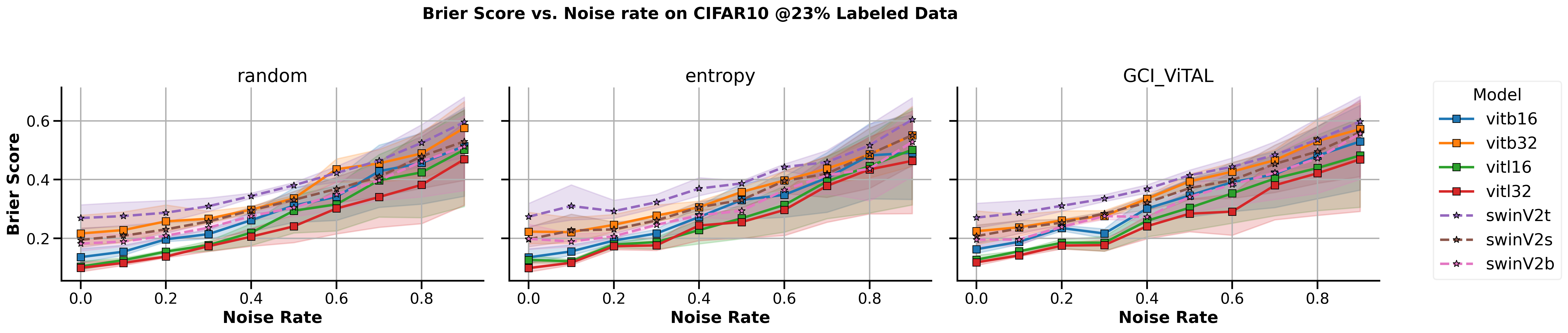}
    \caption{Brier Score vs. Noise Rate on CIFAR10 with 23\% labeled data. Lower values indicate better calibration. Random selection provides more stable calibration than entropy and GCI\_ViTAL.}
\end{figure}

\begin{figure}[htbp]\label{fig:final_brier_vs_noise_rate_cifar10_49}
    \centering
    \includegraphics[width=\textwidth]{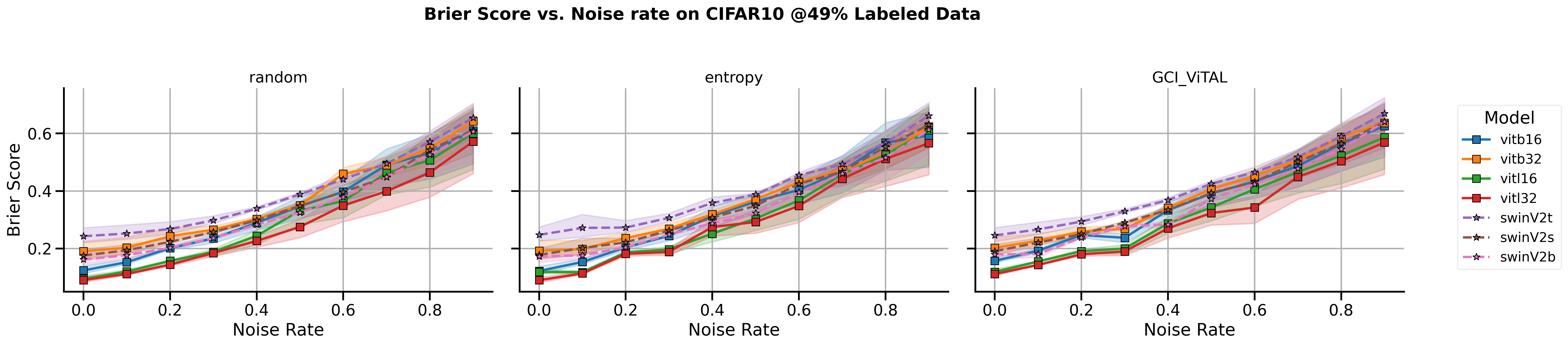} 
    \caption{Brier Score vs. Noise Rate on CIFAR10 with 49\% labeled data. Similar to the 13\% and 23\% label settings, the same trends are maintained, ViTs over SwinV2, and large models over small models when it comes to calibration.}
    
\end{figure}

\subsection*{CIFAR100 Variable Label Noise, Accuracy, and Calibration Curves}
In this section, we provide additional plots in support of our findings on models, accuracy, and calibration.
\begin{figure}[htbp]
    \centering
    \includegraphics[width=\textwidth]{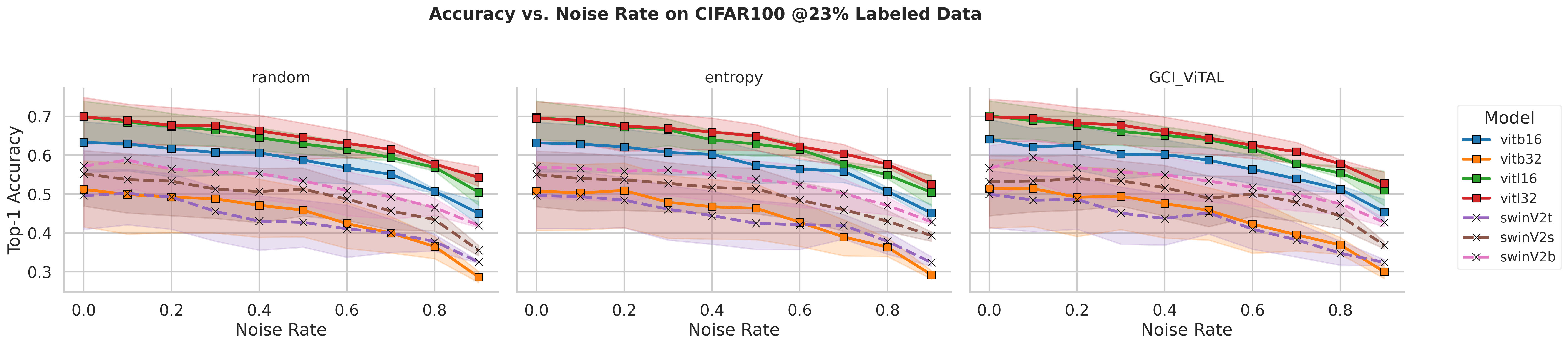} 
    \caption{Accuracy vs. Noise Rate on CIFAR100 with 23\% labeled data. Each subplot shows a different DAL strategy (random, entropy, GCI\_ViTAL) across various Vision Transformer (ViT) and Swin Transformer models.}
\end{figure}

\begin{figure}[htbp]
    \centering
    \includegraphics[width=\textwidth]{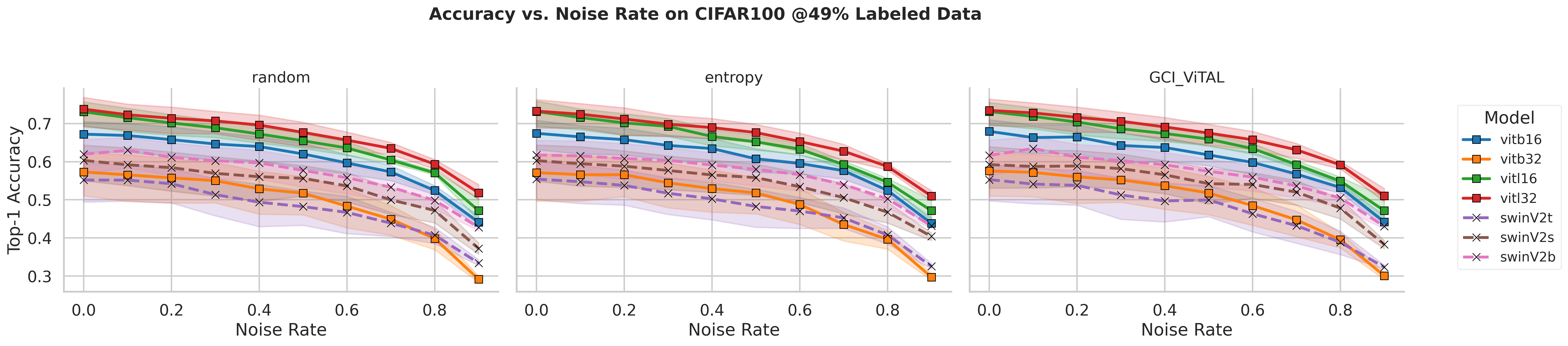}
    \caption{Accuracy vs. Noise Rate on CIFAR100 with 49\% labeled data. ViT models consistently outperform SwinV2 across all noise levels, indicating better robustness to label noise. The GCI\_VITAL active learning strategy maintains higher accuracy compared to random and entropy-based selection, demonstrating its effectiveness in mitigating performance decline due to noise}
\end{figure}

\begin{figure}[htbp]
    \centering
    \includegraphics[width=\textwidth]{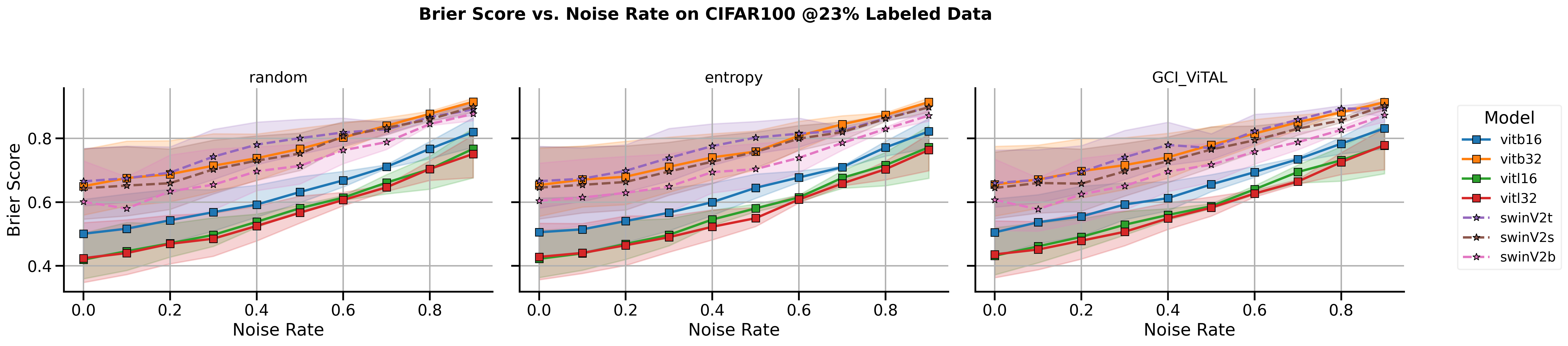}
    \caption{Brier Score vs. Noise Rate on CIFAR100 with 23\% labeled data. The GCI\_VITAL and entropy strategies result in worse Brier Scores compared to random selection, reflective of their accuracy optimizing focus instead of calibration. }
\end{figure}

\begin{figure}[htbp]
    \centering
    \includegraphics[width=\textwidth]{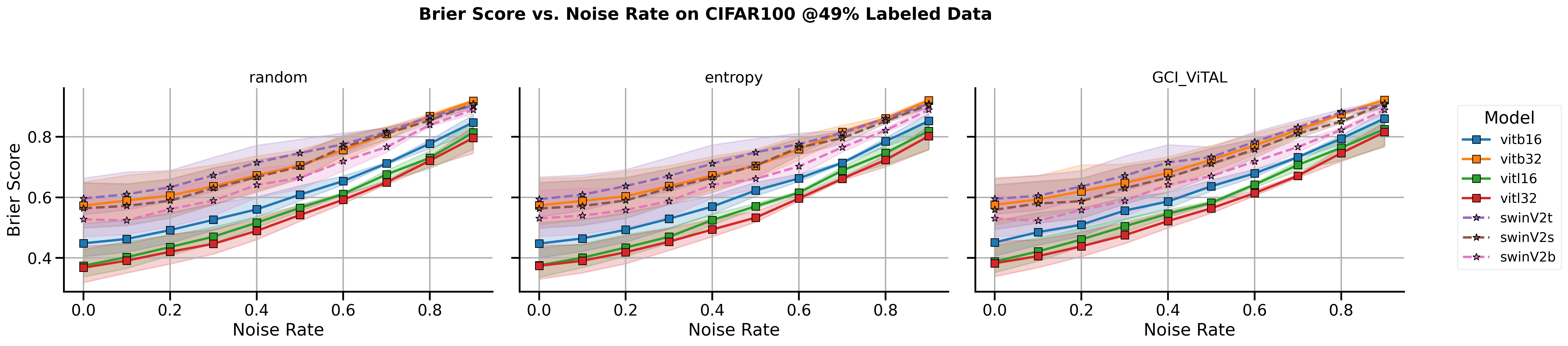}
    \caption{Similar to the trends observed with 13\% and 23\% labeled data, we see consistent patterns across active learning strategies. However, model calibration differences are reduced as label noise increases and the labeled data proportion grows. The gap between small and large models, as well as between ViT and Swin Transformers narrows.}
\end{figure}

\subsection*{Model training Times vs Labeled Data Proportion}
There was little variation between training time and labeled data proportion with all the other experimental variables so this graph summarizes what is expected besides the inefficiency of ViTl16 vs ViTl32.

\begin{figure}[htbp]
    \centering
    \includegraphics[width=0.85\textwidth]{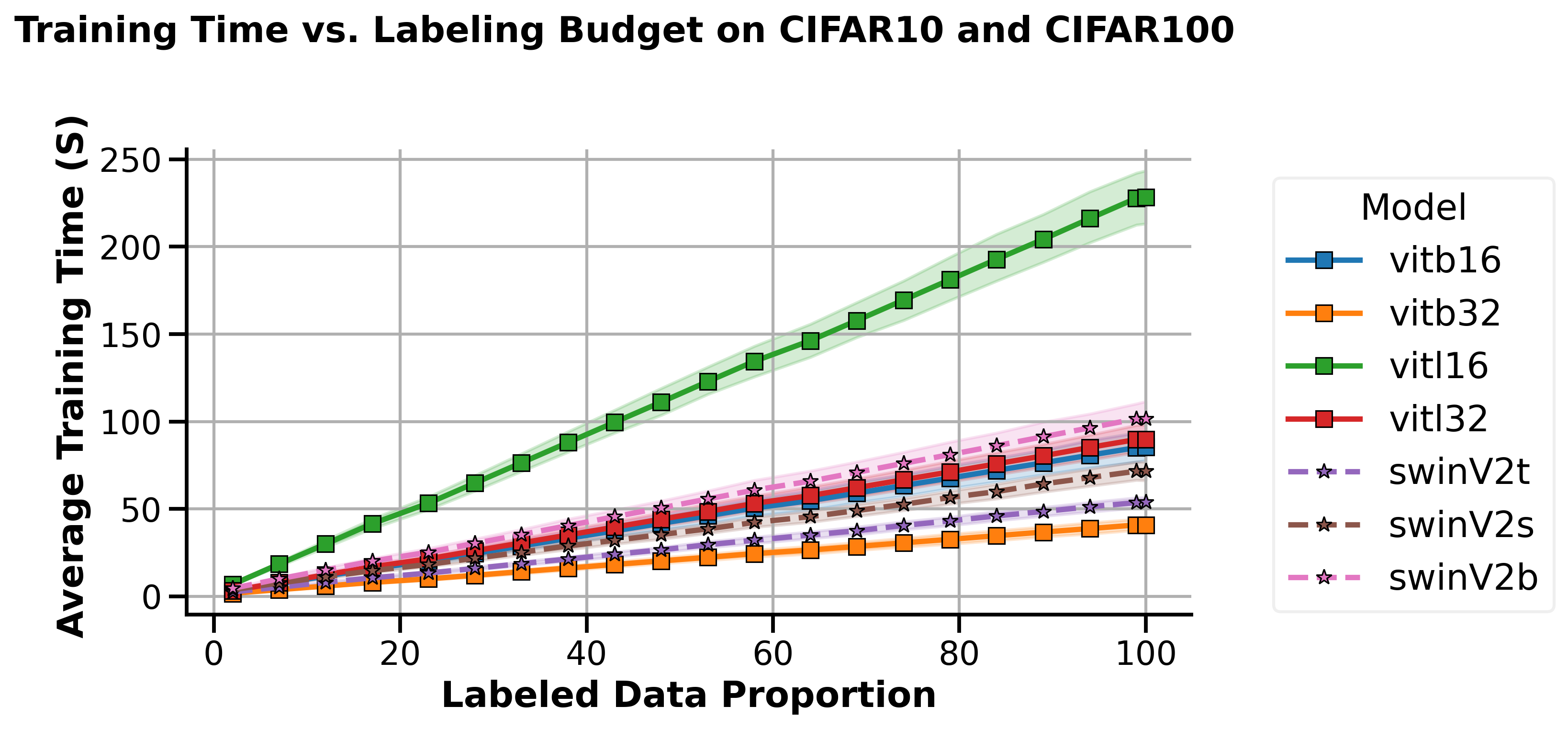} %
    \caption{This graph shows training times for different model configurations against the labeled data proportion averaged over the two datasets, Active Learning strategies, and noise rates. We see a linear trend as expected in the amount of data.}
\end{figure}

\subsection*{Test Accuracy vs Brier Score}
These plots show the interaction between accuracy and efficiency across models and DAL strategies. the results are as expected and do not show any consistent trend that depends on the model choice, this is expected.

\begin{figure}[htbp]
    \centering
    \includegraphics[width=0.95\textwidth]{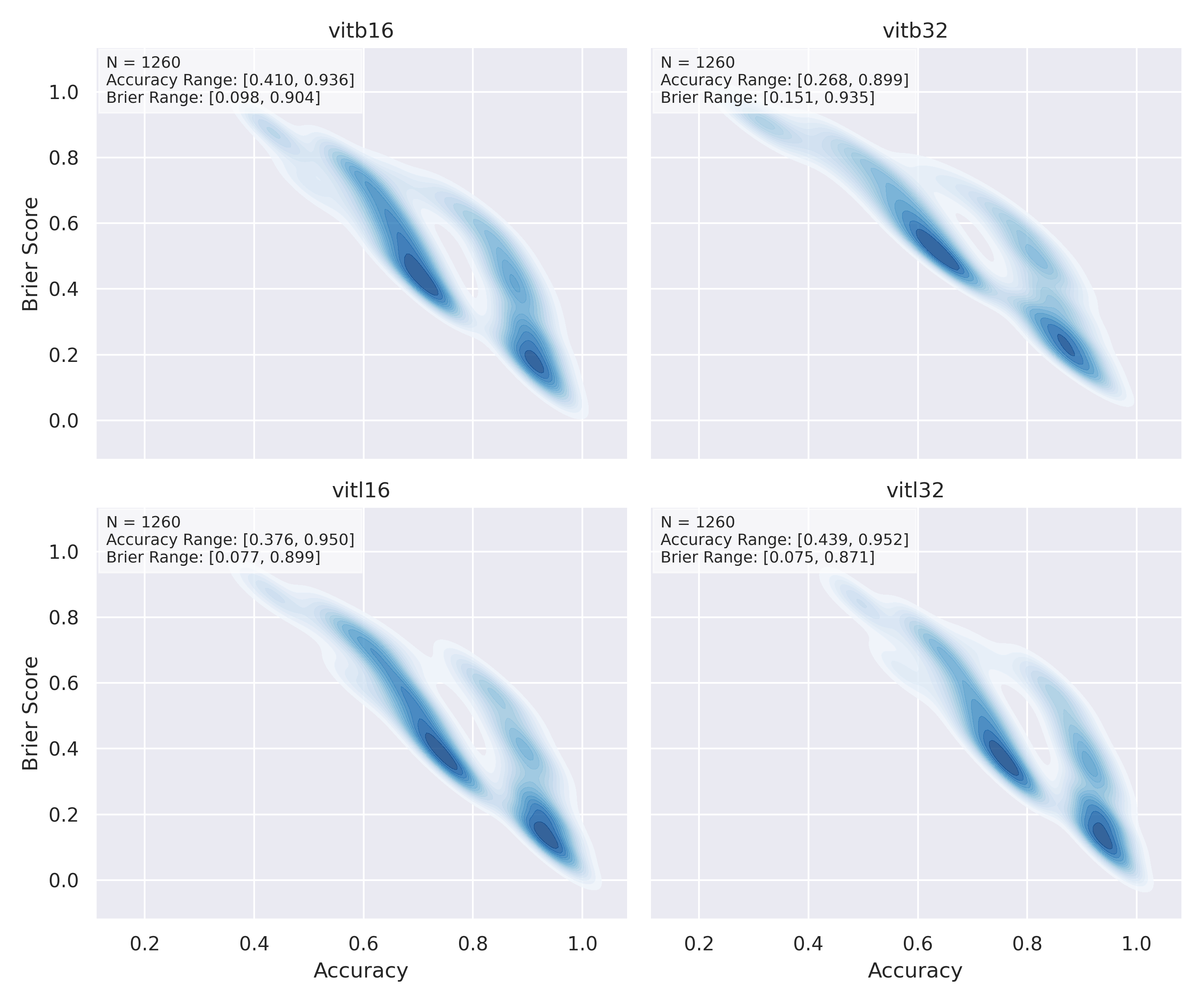}
    \caption{%
    Two-dimensional kernel density estimates (KDEs) of test accuracy vs. Brier score for Vision Transformers (ViT) under 30\% label noise. Each subplot shows a different variant, with darker areas indicating denser run clusters. Lower Brier scores reflect better calibration, while higher accuracy indicates stronger predictive performance. The visible gaps mark regions where model confidence varies despite similar accuracy. 
    }
    \label{fig:acc_brier_vit_swin_fig_kde}
\end{figure}

\begin{figure}[htbp]
    \centering
    \includegraphics[width=\textwidth]{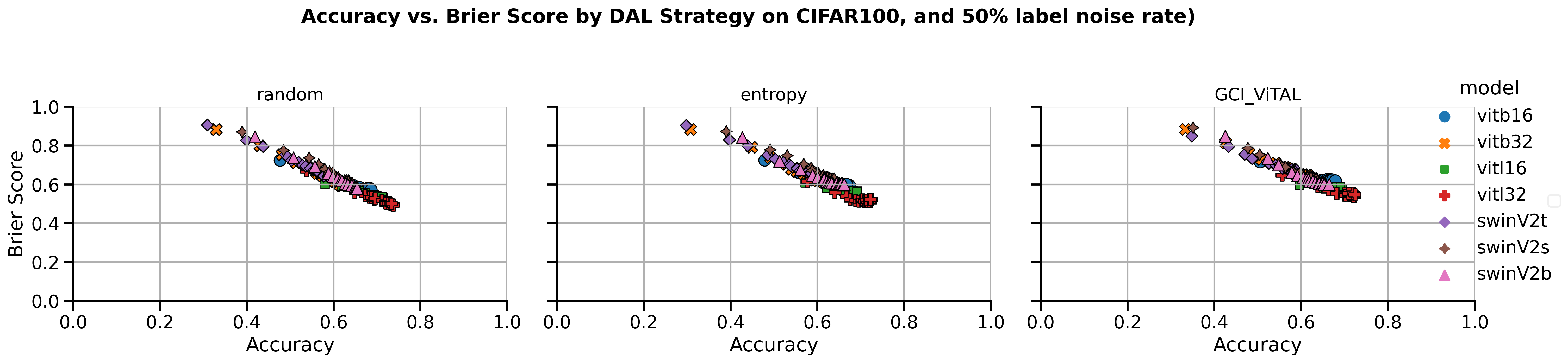} 
    \caption{Accuracy vs. Brier Score by active learning strategy on CIFAR100 at 50\% label noise. At 50\% label noise, ViT models consistently achieve higher accuracy and better calibration than Swin Transformers across all active learning strategies. The accuracy-calibration trade-off remains evident, but ViT models, especially ViTl32 and ViTl16, generally show better calibration, indicating more reliable confidence estimates compared to SwinV2 models.}
    \label{fig:acc_vs_brier_by_strategy_0.5}
\end{figure}

\end{document}